\documentclass[lettersize,journal]{IEEEtran}

\usepackage{amsmath,amssymb,amsfonts}
\usepackage[vlined,ruled,linesnumbered]{algorithm2e}
\usepackage{array}
\usepackage[caption=false,font=normalsize,labelfont=sf,textfont=sf]{subfig}
\usepackage{textcomp}
\usepackage{xcolor}
\usepackage{stfloats}
\usepackage{url}
\usepackage{verbatim}
\usepackage{graphicx}
\usepackage{cite}
\usepackage{amssymb}
\usepackage{mathrsfs}
\usepackage{pbox}
\usepackage{booktabs}
\usepackage{multirow}
\usepackage[numbers,sort&compress]{natbib}
\usepackage{hyperref}

\newtheorem{theorem}{Theorem}

\newtheorem{remark}{Remark}
\newtheorem{lemma}{Lemma}

\newtheorem{assumption}{Assumption}

\graphicspath{{figures/}}

\begin{document}
	
	\title{Concurrent-Learning Based Relative Localization in \\Shape Formation of Robot Swarms}
	
	\author{
		\vskip 1em
		{Jinhu L\"u}, \emph{IEEE Fellow},
		Kunrui Ze, 
		Shuoyu Yue,
		Kexin Liu, 
		Wei Wang, \emph{IEEE Senior Member},
		Guibin Sun
		\thanks{
			This work was supported in part by the National Natural Science Foundation of China under Grants 62373019 and 62141604, in part by the China Postdoctoral Science Foundation under Grant 2023M740185 and in part by the National Key Laboratory of Multi-perch Vehicle Driving Systems under Grant QDXT-NZ-202407-01. 
			\emph{(Corresponding authors: Wei Wang; Guibin Sun.)}}
		\thanks{
			Kunrui Ze, Shuoyu Yue and Guibin Sun are with the School of Automation Science and Electrical Engineering, Beihang University, Beijing 100191, China (e-mail: kr\_ze@buaa.edu.cn; 20351032@buaa.edu.cn; sunguibinx@buaa.edu.cn).
			
			{Jinhu L\"u}, Kexin Liu and Wei Wang are with the School of Automation Science and Electrical Engineering, Beihang University, Beijing 100191, China, and also with the Zhongguancun Laboratory, Beijing 100083, China.
			{Jinhu L\"u} and Wei Wang are also with the Hangzhou Innovation Institute, Beihang University, Hangzhou 310051, China.
			(e-mail: jhlu@iss.ac.cn; kxliu@buaa.edu.cn; w.wang@buaa.edu.cn).}
	}
	
	
	
	\maketitle
	
	\begin{abstract}
		In this article, we address the shape formation problem for massive robot swarms in environments where external localization systems are unavailable. 
		Achieving this task effectively with solely onboard measurements is still scarcely explored and faces some practical challenges.	
		To solve this challenging problem, we propose the following novel results. 
		Firstly, to estimate the relative positions among neighboring robots, a concurrent-learning based estimator is proposed. 
		It relaxes the persistent excitation condition required in the classical ones such as the least-square estimator.  
		Secondly, we introduce a finite-time agreement protocol to determine the shape location. 
		This is achieved by estimating the relative position between each robot and a randomly assigned seed robot. 
		The initial position of the seed one marks the shape location. 
		Thirdly, based on the theoretical results of the relative localization, a novel behavior-based control strategy is devised. 
		This strategy not only enables the adaptive shape formation of large groups of robots but also enhances the observability of inter-robot relative localization. 
		Numerical simulation results are provided to verify the performance of our proposed strategy compared to the state-of-the-art ones. 
		Additionally, outdoor experiments on real robots further demonstrate the practical effectiveness and robustness of our methods. 
	\end{abstract}
	
	\def\abstractname{Note to Practitioners}
	\begin{abstract}
		Shape formation has a broad potential for large groups of robots to execute certain tasks, such as object transport, forest firefighting, and entertainment shows. 
		However, most of the existing approaches rely on external localization infrastructures, rendering them impractical in environments where such systems are not available. 
		To address this issue, this article proposes an integrated strategy that can achieve shape formation for large groups of robots by using local distance and displacement measurements. 
		This strategy consists of three main components.  
		Firstly, a relative localization estimator is introduced to estimate the relative positions among neighboring robots.
		Secondly, a protocol for reaching a consensus on the desired shape's position is proposed. 
		Thirdly, a behavior-based controller is developed to achieve massive shape formation and enhance the observability of relative localization.
		More details of the proposed algorithms and swarm robotic systems are provided in this article. 
	\end{abstract} 
	
	
	\section{Introduction}
	\label{Sec_introduction}
	
	\IEEEPARstart{S}{hape} formation has received significant attention in recent years due to its broad potential across many applications \cite{Alhafnawi2021RAL, Yang2022NMI}. 
	The objective of shape formation is to guide the robot swarms from an initial configuration to form a user-specified shape through local interactions. 
	To achieve this objective, various control methods have been explored, including the ones based on artificial potential \cite{Chu2023TASE}, edge following \cite{Rubenstein2014Science}, mean-shift exploration \cite{Sun2023NC,Zhang2024RAL}, and goal assignment \cite{Wang2020TRO}. 
	However, most of the methods rely on external localization systems such as GPS, 
	to localize robots, which can be restrictive for practical applications \cite{Xie2020TRO}.
	For instance, external localization systems may be impractical or infeasible in some scenarios, such as indoor environment, underwater, or building atrium. 
	Rather than relying on external localization systems, robots equipped solely with onboard sensors are more autonomous. 
	\begin{figure} [!t]
		\centering
		\includegraphics[scale=1]{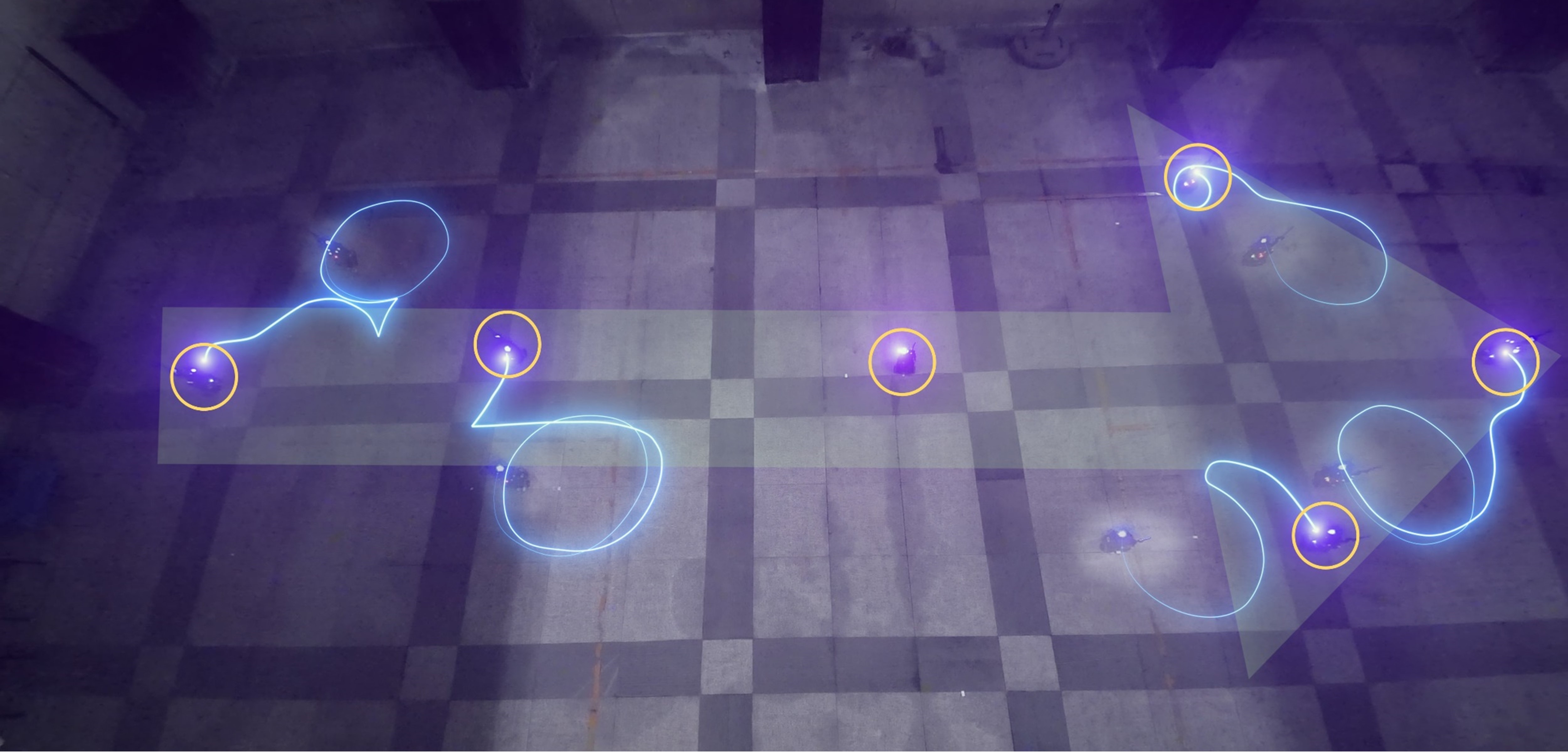}
		\caption{Trajectories of a swarm of six robots forming an arrow shape in an external localization system denied environment. 
			Each robot is equipped with an onboard light. This photo is obtained by long-exposure photography. }
		\label{arrow}
	\end{figure}
	
	In this article, we aim to investigate the problem of relative localization in shape formation using only onboard measurements, as shown in Fig.~\ref{arrow}. 
	To handle this problem, the existing results can be roughly classified into the following two categories. 
	The first class is based on the \emph{rigidity} conditions. 
	These methods usually require an assumption that the measurement topology among robots satisfy rigidity conditions like distance rigidity \cite{Shen2023TCNS, Mehdifar2020Auto} or bearing rigidity \cite{Zhao2016Auto, Zhao2019TAC}. 
	The second class, which has also attracted research attention, is rigidity-free methods.  
	To that end, researchers incorporate the \emph{persistent excitation} (PE) characteristics into the control command, such as noise \cite{Liu2023Auto, Xie2019TCNS} or time-varying reference \cite{Tang2022Auto, Xie2020TRO}, to relax rigidity conditions on the measurement topology. 
	Although rigidity and PE conditions are necessary in some collaborative tasks, they may not be essential in the context of massive shape formation. 
	More significantly, imposing these conditions may bring a series of technical challenges, as outlined below.
	
	Firstly, the rigidity condition is the foundation of the global convergence property of most existing results \cite{Shen2023TCNS, Mehdifar2020Auto, Zhao2016Auto, Zhao2019TAC}.
	The rigidity condition indicates that any two neighboring robots must involve onboard measurements and the measurement topology for the entire robot swarms should be rigid. 
	However, the two robots that are topological neighbors may be far apart physically, which brings difficulties in measurement and communication between them.
	Furthermore, the rigidity condition across the entire topology is also difficult to guarantee, especially for large groups of robots.
	
	Secondly, to relax the rigidity condition, researchers introduce the PE condition based localization estimator.
	In practice, the most common way to fulfill the PE condition is embedding noise or time-varying reference into the control commands of individual robots \cite{Liu2023Auto,Xie2019TCNS}.
	However, these PE characteristics may propagate within robot swarms due to interactions between robots.
	This can result in non-smooth motion trajectories for robot swarms, leading to deteriorated control performance in real-world applications.

	Motivated by these challenges, we focus on the shape formation control for large groups of robots without PE condition. 
	The contributions of this article are summarized below. 
	
	1) To achieve inter-robot relative localization without the PE condition, a concurrent-learning based relative position estimator is firstly introduced. 
	Compared with the existing results \cite{Xie2019TCNS, Liu2023Auto, Tang2022Auto}, both current and historical measurement data are utilized in the estimator. 
	As a result, it is not necessary to incorporate PE characteristics into the control command, resulting in smoother motion trajectories for the robots.
	Theoretical analysis results are provided to guarantee the convergence of the estimation error.
	Furthermore, we provide the optimal strategy for relative motion and data collection between robots. We demonstrate that the proposed estimator achieves the largest convergence rate with the optimal strategy.
	
	2) To achieve the shape localization agreement for the entire swarms in the absence of global coordinates, the location of the shape center is selected as the initial position of a randomly assigned seed robot.
	Note that the position of the seed robot is unavailable and irrelevant.
	By transforming the global-coordinate based shape agreement problem \cite{Sun2023NC} into a relative position estimation problem, we propose a finite-time consensus-based agreement protocol.
	This approach eliminates the need for the rigidity condition on the measurement topology \cite{Shen2023TCNS, Mehdifar2020Auto, Zhao2016Auto, Zhao2019TAC} to achieve shape localization agreement. 
	Theoretical analysis results are also provided to guarantee the convergence of the proposed method. 
	
	3) For the shape formation control of the robot swarms, we design a novel behavior-based control strategy which is based on the theoretical insights from relative localization. 
	This strategy enables shape formation in large groups of robots in the absence of external localization system. 
	Additionally, it enhances the observability of relative localization among robots.
	Extensive simulation results are provided to verify the effectiveness of the proposed methods.
	Moreover, compared with the existing swarm shape formation methods based on relative localization \cite{Xie2020TRO, Xie2023TRO, Xu2022TRO}, outdoor experiments with a number of robots up to six are also provided to verify the effectiveness, stability, and robustness of the proposed method. 
	
	The rest of this article is organized as follows. 
	Section \ref{Sec_relatework} provides the related works. 
	Section \ref{Sec_overview} introduces an overview of the entire system, adaptive relative localization method and behavior-based shape formation control strategy are proposed in Section \ref{Sec_localization} and Section \ref{Sec_shapeformation}. 
	Simulation and experimental results are discussed in Section \ref{Sec_simulation} and Section \ref{Sec_experiment}. 
	Finally, Section \ref{Sec_conclusion} concludes this article.
	
	\section{Related Works}
	\label{Sec_relatework}
	
	\subsection{Distributed Shape Formation}
	\label{Sec_formation_work}
	Traditional researches on swarm formation control rely on consensus and graph theory \cite{Wang2014Auto, Ze2023TIE, Zhao2020Auto}. 
	In these methods, each robot achieves shape formation by tracking a predefined desired reference in a distributed manner. 
	However, the neighboring robots defined in the desired formation may be far apart physically, which is unreasonable in real applications.
	
	One class of shape formation strategies that are widely studied is based on goal assignment. 
	In these methods, each robot is dynamically assigned a unique goal location using decentralized approaches such as distributed auction \cite{Morgan2016IJRR} and task-swapping algorithm \cite{Wang2020TRO}.
	Once the robots are assigned unique goal locations, the subsequent task is simply to guide the robots to reach their respective goal locations. 
	Many existing results, such as distributed model predictive control \cite{Morgan2016IJRR}, optimal reciprocal collision avoidance \cite{Van2011VO} and planning based methods \cite{Zhou2022SR}, can accomplish this task effectively.
	
	The other class of strategies are based on local
	behavior \cite{Chen2014JRSI, Krieger2000Nature, Gelblum2015NC}.
	Inspired by natural processes like embryogenesis, the method in \cite{Slavkov2018SR} can spontaneously generate swarm shapes with a reaction-diffusion network. 
	Similarly, a random walk based method is proposed in \cite{Alhafnawi2021RAL}.
	These methods exhibit significant randomness in both the shape formed and the robots' motion trajectories. 
	Another common assignment-free methods for shape formation are based on artificial potential fields (APF). However, APF-based methods often encounter a well-known \emph{locally minimum} problem.
	To address this issue, an information integration and feedback based method is proposed in \cite{Chu2023TASE}. 
	However, relying solely on APF cannot effectively avoid the deadlock phenomena.
	In our previous work \cite{Sun2023NC}, the idea of mean-shift exploration is adopted in the shape formation of robot swarms, which significantly improves the efficiency of shape formation, especially for large groups of robots.
	
	All these methods are based on the robots' global coordinates. 
	However, one can see that many simple organisms in nature can achieve complex shapes with only basic measurement information rather than global coordinates. 
	Consequently, the development of shape formation techniques for robot swarms using only local measurements remains significant.
	
	\subsection{Relative Localization Estimation}
	\label{Sec_localization_work}
	
	Relative localization between robots is a fundamental problem in swarm robot research. 
	Different from traditional anchor-based localization methods \cite{Patwari2005TSPM, Nguyen2016TSP}, we focus on the relative localization strategy which does not rely on external anchors \cite{wang2018IJSAC}.
	In recent years, visual-measurement based relative localization schemes have received significant attention. 
	The work in \cite{Zhao2022TRO} proposes a fisher information matrix (FIM) based localization-enhanced method with bearing measurement. 
	In \cite{wang2023IJSAC} and \cite{wang2023RAL}, partially observations based framework is proposed to estimate the relative poses within a robotic swarm.
	To handle the limited camera's field of view, an omnidirectional cameras based simultaneous localization and mapping system named $D^2$SLAM is proposed in \cite{xu2024TRO}. 
	However, the vision-based relative localization methods may lead to high computational resource consumption \cite{Xie2023TRO}.
	
	In contrast, thanks to the availability of low-cost ultra wide band (UWB) ranging sensors, distance based localization schemes are also widely explored. 
	A common approach is to install multiple UWB modules on the robot.
	In the lastest results \cite{shalaby2024TRO} and \cite{cano2023TRO}, the relationship between the motion trajectory and localization performance of robot is studied.
	However, the use of multiple UWB sensors may increase the structural complexity of the robot.
	In \cite{Xie2019TCNS}, a distance and displacement measurement based relative position estimator is proposed, requiring only one UWB sensor. 
	However, this estimator only converges when the relative motion trajectory between robots satisfies the PE condition. 
	As a result, the periodic noise signal is added to the control command, which is similar to \cite{Liu2023Auto}. 
	In \cite{Xie2023TRO}, based on the FIM theory, the authors analyze how the relative motion between two robots can enhance relative localization. 
	In the latest result \cite{Liu2023RAL}, the PE condition is satisfied through the rotation of a UWB module on each robot. 
	Nevertheless, the design of the turntable adds complexity to the robot's structure.
	
	In summary, most existing works still rely on the PE conditions in the relative localization task. 
	However, it is worth noting that with the careful use of both the current and historical measurement data, this condition can be relaxed.
	As a result, the smoothness of the robots' motion trajectories can be directly improved. 
	
	\subsection{Local Measurement Based Shape Formation}
	\label{Sec_localization_formation_work}
	To achieve shape formation tasks with only local measurements, researchers in control and robotics have explored various strategies.
	One common strategy is based on a trilateral localization algorithm. 
	In \cite{Rubenstein2014Science}, a distance measurement based shape formation strategy is proposed. 
	In this work, each robot localizes itself by measuring distance with at least four localized robots. 
	However, this method requires the placement of anchor robots and only edge robots can move to form shapes, resulting in lower efficiency.
	Similarly, multiple ranging sensors are used for inter-robot localization in \cite{Guler2020TCST}. 
	In \cite{Pratissoli2023NC}, the elastic link is adopted to enhance localization for error-prone individuals. 
	However, these methods rely on external facilities, making practical implementation challenging.
	
	To overcome the above shortcomings, researchers have developed shape formation control methods that do not require external anchors. 
	In \cite{Shen2023TCNS}, a distance-based adaptive formation control method is proposed based on the graph rigidity theory. 
	Similarly, a bearing rigidity based shape formation control method is introduced in \cite{Zhao2019TAC}. 
	However, these methods necessitate that the measurement topology between robots satisfies specific conditions, which can be impractical in real-world implementations.
	To address these issues, adaptive parameter estimation is used for relative localization in works such as \cite{Xie2020TRO, Tang2022Auto}. 
	Specially, when the relative motion trajectory between two robots satisfies the PE condition, the convergence of these methods can also be guaranteed under general connected measurement topology. 
	However, PE-based methods may be ineffective for tasks such as static shape formation or linear cruising formation where the PE condition is not satisfied.
	
	In summary, while the existing approaches offer promising solutions for local measurement based swarm shape formation control, they all have limitations. Currently, there is still no method that achieves efficient shape formation control for large groups of robots without relying on the PE condition. The distributed shape formation problem with solely onboard measurements remains a challenge yet to be fully addressed.
	
	\section{Preliminaries and System Overview}
	\label{Sec_overview}
	
	\subsection{Notions}
	
	\begin{table}[!t]
		\caption{List of main symbols used in this article}
		\centering
		\begin{tabular}{ll}
			\hline\hline \\[-3mm]
			\multicolumn{1}{c}{Symbols} & \multicolumn{1}{c}{Description} \\ \hline \\[-2mm]
			$ t_{k} $  & \pbox{10cm}{UWB \& IO measurement time instant for robot $i$. } \\ [1.3ex] 
			$ p_{i}(t_{k}) $  & \pbox{10cm}{Position of the robot $i$. } \\ [1.3ex] 
			$ u_{i}(t_{k}) $  & \pbox{10cm}{Displacement of robot $i$ from time instant $t_{k}$ to $t_{k+1}$. } \\ [1.3ex]
			$ z_{i}(t_{k}) $  & \pbox{10cm}{Displacement of robot $i$ from initial time instant $t_{0}$ to $t_{k}$. } \\ [1.3ex]
			$ d_{ij}(t_{k}) $ & \pbox{10cm}{Distance between robots $i$ and $j$. } \\ [1.3ex]
			$ t_{{\rm c},m} $ & \pbox{10cm}{Starting time instant of the $m$-th measurement data collection. } \\ [1.3ex]
			$ y_{ij} $ & \pbox{10cm}{Auxiliary signal which records innovation. } \\ [1.3ex]
			$ \hat{p}_{ij} $ & \pbox{10cm}{Estimate of real-time relative position between robots $i$ and $j$. } \\ [1.3ex]
			$ \hat{q}_{i,0} $ & \pbox{10cm}{Estimate of initial relative position between robot $i$ and the \\ seed robot $0$. } \\ [1.3ex]
			$ v_{i} $ & \pbox{10cm}{Velocity control command for robot $i$. } \\ [1.3ex]
			\hline\hline \\[-4mm]
		\end{tabular} \label{Table1}
	\end{table}

	\begin{figure*}[!t]\centering
		\includegraphics[scale=1.066]{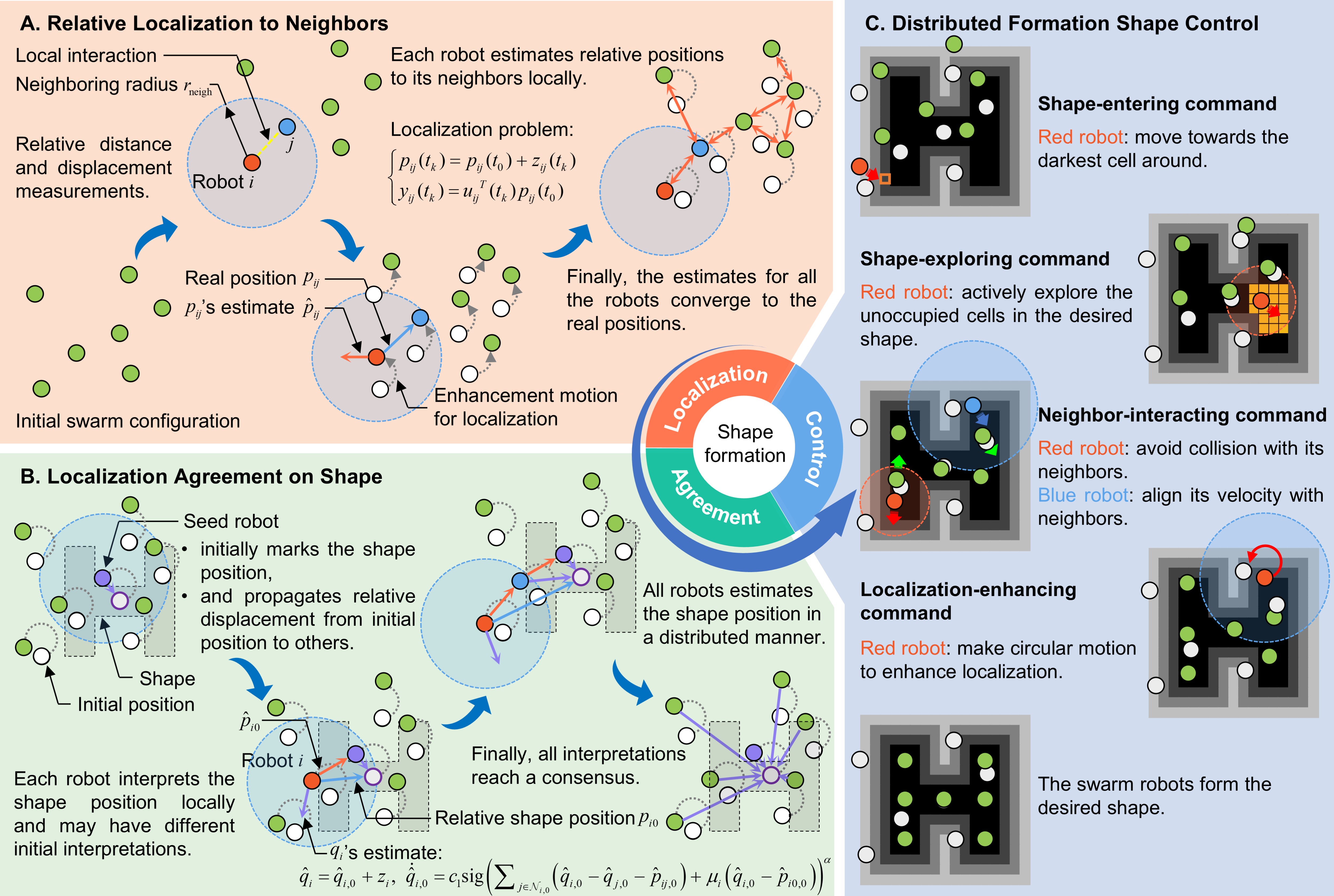}
		\caption{An illustration of the proposed shape formation strategy. \textbf{A}. Each robot in the swarms moves with the localization enhance control law and collects data to localize its neighboring robots. \textbf{B}. Each robot estimate the relative position to the initial position of the seed robot with the finite-time consensus method. \textbf{C}. Behavior-based localization enhance shape formation control scheme.}\centering
		\label{overview}
		\vspace{-0.5cm}
	\end{figure*}
	
	Consider a group of $n+1$ mobile robots in $\mathbb{R}^2$ and $n\geq 2$. 
	Let $p_i \in \mathbb{R}^2$ be the position of robot $i=0,1,\ldots,n$ in an inertial frame. 
	The dynamic model of each robot is considered as $\dot p_i=v_i$, where $v_i$ is the velocity command to be designed.  
	The velocity command $v_{i}$ satisfies $\lVert v_{i} \rVert \leq v_{\max}$, where $v_{\max}$ is the maximum velocity of each robot. 
	The interaction network among the robots is described by an undirected graph $\mathcal{G}=(\mathcal{V},\mathcal{E})$, which is composed of a vertex set $\mathcal{V}=\{0,1,\ldots,n\}$ and an edge set $\mathcal{E}\subseteq \mathcal{V}\times\mathcal{V}$.
	If $(i,j)\in \mathcal{E}$, the two robots can receive information from each other and robot $j$ is a neighbor of robot $i$. 
	The neighbor set of robot $i$ is denoted as $\mathcal{N}_i=\{j\in \mathcal{V}: (i,j)\in \mathcal{E}\}$. The detailed definition of the neighbor set $\mathcal{N}_{i}$ can be found in \ref{Sec_agreement}. 
	For ease of reading, we have listed the main symbols that appear in this article in Table \ref{Table1}.
	
	\subsection{System Overview}
	The objective of each robot is to achieve shape formation for the entire robot swarms through local distance and displacement measurements.  
	To achieve this, we design an integration strategy including relative position estimation, shape localization agreement, and shape formation control. 
	The overall system architecture is illustrated in Fig.~\ref{overview}. 

	The first step involves designing a concurrent-learning based estimator to estimate relative positions to neighboring robots (see Fig.~\ref{overview}A and Section \ref{Sec_localization}). This article enables each robot to achieve relative localization to its neighbors solely through inter-robot distance and displacement measurements which can be obtained using some low-cost sensors such as UWB and inertial odometry (IO), respectively. 
	The proposed distributed adaptive estimators for relative position estimation among neighboring robots are based on the concept of concurrent-learning, i.e. the robots leverage historical and present measurement data to alleviate the need for the PE condition.

	The second step is to design an agreement protocol to locate the desired shape which is marked by the initial position of the seed robot (see Fig.~\ref{overview}B and Section \ref{Sec_agreement}).
	The seed robot plays a stubborn role by insisting on propagating the relative displacement from its initial position (i.e., shape position) to its non-seed neighbors. 
	These non-seed robots utilize this propagated relative displacement to interpret the shape position locally through the proposed agreement protocol, simultaneously propagating their interpretations to other non-seed robots. 
	In this way, the interpretations of non-seed robots gradually converge to the seed one so that all the robots reach a consensus on the localization of the desired shape.

	The final step is to devise a behavior-based control law to achieve the shape formation task and, meanwhile, enhance the localization (see Fig.~\ref{overview}C and Section \ref{control}).
	Once all the robots reach an agreement on the localization of the desired shape, each robot can form the desired shape driven by the proposed behavior-based control law (see Section \ref{Sec_shapeformation}). 
	Specifically, the behavior-based control law consists of four velocity commands $v_i^{\rm ent}$, $v_i^{\rm exp}$, $v_i^{\rm int}$, and $v_i^{\rm enh}$. The four commands representthe shape-entering, shape-exploring, neighbor-interacting, and localization-enhancing behavior motions, respectively.
	
	\section{Adaptive Relative Localization}
	\label{Sec_localization} 
	
	This section addresses the problem of achieving relative localization among adjacent robots, through the use of onboard measurements, such as inter-robot distance and displacement. 
	The fundamental idea is to formulate the relative localization problem as a parameter estimation problem, and then design a concurrent-learning based estimator to estimate relative positions to neighboring robots. 
	The details are as follows. 
	
	\subsection{Relative Localization Problem} 
	\label{Sec_problem} 
	
	For each robot $i\in \mathcal{V}$ and $j\in \mathcal{N}_i$, denote the relative position between robots $i$ and $j$ at time $t_k=k\Delta t$ as $p_{ij}(t_k)=p_{i}(t_k)-p_{j}(t_k)$,
	where $k\in \mathbb{N}$ is the number of samples, and $\Delta t$ is the sampling interval. 
	Here, $p_{ij}(t_k)$ is related to its initial relative position $p_{ij}(t_0)$ and satisfies
	\begin{align}
		p_{ij}(t_k)=p_{ij}(t_0)+z_{ij}(t_k)
		\label{Equ_initial}
	\end{align}
	where $z_{ij}(t_{k})$ is defined as $z_{ij}(t_{k}) = z_{i}(t_{k})-z_{j}(t_{k})$, $z_{i}(t_k)$ is the displacement measured by IO of robot $i$ from $t_0 = 0$ to $t_k$, i.e. $z_{i}(t_{k})=\int_{t_{0}}^{t_{k}} v_{i}(t) d t$. 
	Note that $z_{j}(t_{k})$ is obtainable through the wireless communication between robots $i$ and $j$, then $z_{ij}(t_{k})$ is also available.
	However, the initial relative position $p_{ij}(t_0)$ of the robot cannot be measured solely through sensors such as UWB and IO.
	Then we can formulate the relative localization problem, i.e., the determination of $p_{ij}(t_{k})$, as an estimation problem of parameter $p_{ij}(t_{0})$.

	According to the cosine theorem, the relative distance, relative displacement and relative position between robot $i$ and robot $j$ satisfy
	\begin{align}
		\label{Equ_relative}
		u_{ij}^T(t_{k})p_{ij}(t_{k}) = \frac{1}{2}\left( d_{ij}^2(t_{k+1}) - d_{ij}^2(t_{k}) - u_{ij}^2(t_{k})\right)
	\end{align}
	where $d_{ij}(t_{k})$ represents the relative distance between robots $i$ and $j$ at time instant $t_{k}$ and is defined as $d_{ij}(t_{k}) = \lVert p_{ij}(t_{k}) \rVert$. 
	$u_{ij}(t_{k})$ is defined as $u_{ij}(t_{k}) = u_{i}(t_{k})-u_{j}(t_{k})$, where $u_{i}(t_k) = \int_{t_{k}}^{t_{k+1}} v_{i}(t) d t$ is the displacement measured by IO of robot $i$ from $t_k$ to $t_{k+1}$. Since $u_{i}$ and $u_{j}$ can be measured by IO, $u_{ij}$ can be directly obtained through the wireless interaction between robots $i$ and $j$. 
	For a better understanding of \eqref{Equ_relative}, an illustrative example is introduced in Fig. \ref{data_collect}. As shown in Fig. \ref{data_collect}A, it can be seen that the relative displacement $u_{ij}(t_{k})$ and the relative positions between the two robots at $t_{k}$ and $t_{k+1}$ form a triangle.
	
	Define an auxiliary signal $y_{ij}(t_{k})$ as,
	\begin{align}
		\label{Equ_problem}
		y_{ij}(t_{k})&= u_{ij}^T(t_{k})p_{ij}(t_{0})
		\\
		\notag
		&=u_{ij}^T(t_{k})p_{ij}(t_{k})-u_{ij}^T(t_{k})z_{ij}(t_{k}) 
		\\
		\notag
		&= \frac{1}{2} \left( d_{ij}^2(t_{k+1}) - d_{ij}^2(t_{k}) - u_{ij}^2(t_{k}) \right) - u_{ij}^T(t_{k})z_{ij}(t_{k}).
	\end{align}
	One can see that $y_{ij}(t_{k})$ is available, since both $u_{ij}$ and $z_{ij}$ are directly obtainable. 
	With the fact that both $y_{ij}(t_{k})$ and $u_{ij}(t_{k})$ can be measured, the adaptive parameter update law can be designed to estimate the initial relative position $p_{ij}(t_{0})$. As a result, the relative localization task can be achieved. 

	Denote $\hat{p}_{ij,0}(t_k)$ as an estimation of the initial relative position $p_{ij}(t_0)$ and $\tilde{p}_{ij,0}(t_k)=\hat{p}_{ij,0}(t_k)-p_{ij}(t_0)$ as the estimation error.
	We can define the innovation $\epsilon_{ij}(t_{k})$ as
	\begin{align}
		\epsilon_{ij}(t_{k}) =u_{ij}^T(t_{k})\hat{p}_{ij,0}(t_{k}) - y_{ij}(t_{k}) = u_{ij}^T(t_{k})\tilde{p}_{ij,0}(t_{k}). 
		\label{Equ_error}
	\end{align}
	The innovation $\epsilon_{ij}(t_{k})$ will be used later to design the concurrent-learning based estimator in Section~\ref{Sec_observer}. 
	Based on the estimation problem modeled as (\ref{Equ_problem}), methods proposed in \cite{Xie2020TRO}, \cite{Liu2023Auto} and \cite{Liu2023RAL} attempt to estimate the initial relative position $p_{ij}(t_{0})$.
	However, all these existing results require that relative displacement $u_{ij}(t)$ satisfies the PE condition. 
	To satisfy this condition, the persistent noise or periodic signal is adopted in the control command of the robot, which is not conducive to the expansion of massive robot swarms.
	\subsection{Data Collection Strategy} 
	\label{Sec_collection} 
	
	\begin{figure}[!t]\centering
		\includegraphics[scale=1.066]{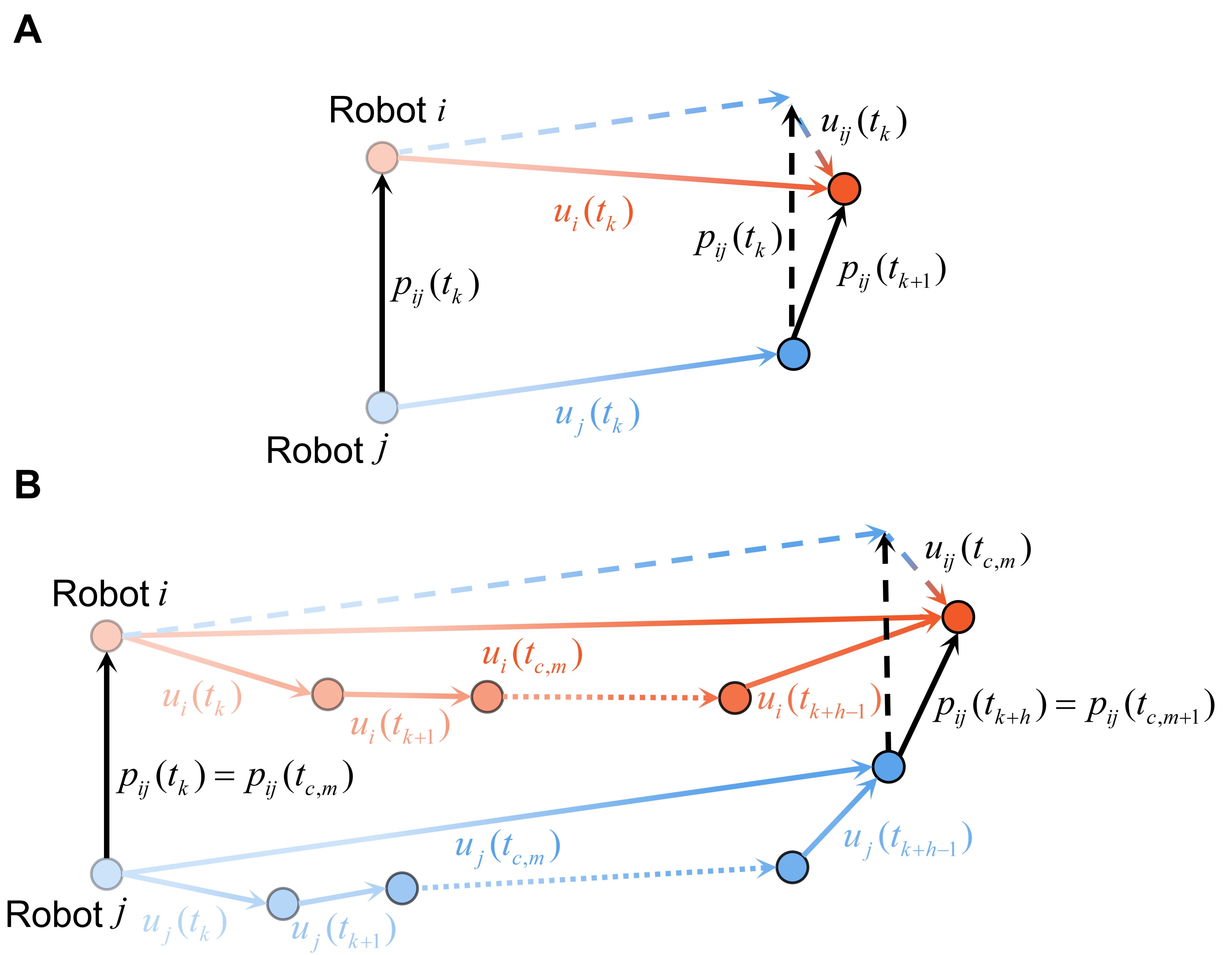}
		\caption{Geometric relationship between the displacements and distance measurements of the two robots. \textbf{A}. One sampling interval $\Delta t$ case. \textbf{B}. A sketch for $h$ sampling interval $\Delta t$ case}\centering
		\label{data_collect}
		\vspace{-0.5cm}
	\end{figure}
	
	In the next section, how to collect varables $d_{ij}(t_k)$, $d_{ij}(t_{k+1})$ and $u_{ij}(t_k)$ to determine $p_{ij}(t_0)$ will be presented. 
	However, for the real IO sensor, if the sampling interval $\Delta t$ is too small, the measurements of $u_{ij}(t_k)$ may be drowned in noise in practice. 
	To address this issue, we propose the following data collection strategy.
	
	Whenever two robots start localization, a data collection takes place between them. 
	As shown in Fig. \ref{data_collect}B, assuming that a data collection starts at time instant $t_{k}$, then it will end at $t_{k+h}= t_{k} + h\Delta t$.
	For biref, we denote $t_{{\rm c}, m}$ and $t_{{\rm c}, m+1}$ as the starting and ending time of the $m$-th ($m=1,\ldots,\varsigma$) data collection. 
	Note that $\varsigma$ is the data collection number and $t_{{\rm c}, m+1}$ is also the starting time of the $(m+1)$-th data collection.
	The interval between the two consecutive collections is $h\Delta t, h\in \mathbb{N}$, i.e., $t_{{\rm c},m+1}-t_{{\rm c},m}=h\Delta t$.
	The whole data collection process begins at $t_{{\rm c}, 1}$ and terminates at $t_{{\rm c}, \varsigma}$.
	The heuristic values of $h$ and $\varsigma$ will be discussed later in Section \ref{Sec_motion_design}. 
	In this way, the innovation $\epsilon_{ij}(t_{k})$ in \eqref{Equ_error} can be rewritten as 
	\begin{align}
		\label{Equ_error_m}
		\epsilon_{ij}(t_{{\rm c},m}) =& u_{ij}^T(t_{{\rm c},m})\hat{p}_{ij,0}(t_{k}) - y_{ij}(t_{{\rm c},m})
		\\
		\notag
		y_{ij}(t_{{\rm c},m})= 
		&\frac{1}{2}\left( d_{ij}^2(t_{c,m + 1}) - d_{ij}^2(t_{{\rm c},m}) - u_{ij}^2(t_{{\rm c},m}) \right)
		\\
		\notag
		&- u_{ij}^T(t_{{\rm c},m})z_{ij}(t_{{\rm c},m})
	\end{align}
	where $z_{ij}(t_{{\rm c},m})=\sum_{\iota=1}^{m}u_{ij}(t_{{\rm c},\iota})$. 
	The innovation $\epsilon_{ij}(t_{{\rm c},m})$ will be used later in Section~\ref{Sec_observer}. Note that the tandem data collection strategy is helpful for increasing the modulus of the relative displacement $u_{ij}(t_{{\rm c},m})$, which is beneficial for reducing relative error in noisy measurements. This trick helps to improve estimation accuracy in real applications.
	
	\subsection{Concurrent-Learning based Estimator} 
	\label{Sec_observer} 
	
	The concurrent-learning based estimator $\hat{p}_{ij,0}(t_k)$, which is used to estimate $p_{ij}(t_{0})$, can be designed as
	\begin{subequations}
		\label{Equ_observer}
		\begin{align}
			\hat{p}_{ij,0}(t) = & \hat{p}_{ij,0}(t_{k}), \quad t_{k} < t \leq t_{k+1} \\
			\hat{p}_{ij,0}(t_{k+1}) = & \hat{p}_{ij,0}(t_{k}) - \eta \sum_{m=1}^{\varsigma } u_{ij}^T(t_{{\rm c},m})\epsilon_{ij}(t_{{\rm c},m}) \\
			\notag
			&- \eta u_{ij}^T(t_{k})\epsilon_{ij}(t_{k})
		\end{align}
	\end{subequations}
	with learning rate $\eta$
	\begin{align}
		\eta = \frac{\lambda_{\min}(S_{ij})}{( \lambda_{\max}(U_{ij}(t_{k})) + \lambda_{\max}(S_{ij}))^2}
		\label{Equ_step}
	\end{align}
	where matrices $U_{ij}(t_{k})$ and $S_{ij}$ are defined as $U_{ij}(t_{k}) = u_{ij}(t_{k}) u_{ij}^T(t_{k})\in \mathbb{R}^{2\times2}$ and $S_{ij}=R_{ij}R_{ij}^T\in \mathbb{R}^{2\times2}$, respectively. Here, $R_{ij}=[u_{ij}(t_{c,1}),...,u_{ij}(t_{c,\varsigma})]$. 
	Note that (\ref{Equ_observer}) represents the updates at time instant $t_{k+1}$ and employs both the present measurements $d_{ij}(t_{k+1})$, $d_{ij}(t_{k})$, $u_{ij}(t_{k})$ and the historical measurements $d_{ij}(t_{{\rm c},m})$, $d_{ij}(t_{c,m+1})$, $u_{ij}(t_{{\rm c},m})$ in its calculations. All these measurements are available at $t_{k+1}$. 
	
	To relax the PE condition, a concurrent-learning term $\eta \sum_{m=1}^{\varsigma } u_{ij}^T(t_{{\rm c},m})\epsilon_{ij}(t_{{\rm c},m})$ which records the historical measurement information is designed in the estimator (\ref{Equ_observer}). As a result, the persistent noise velocity command can be replaced by an impermanent localization enhance velocity command and the motion trajectories of robots will be smoother. 
	Based on the initial state estimator (\ref{Equ_observer}), the real-time relative position estimator $\hat{p}_{ij}(t)$ can be designed as
	\begin{align}
		\hat{p}_{ij}(t) = \hat{p}_{ij,0}(t) + z_{ij}(t).
		\label{Equ_location}
	\end{align}
	Recall that $z_{ij}(t)$ is the relative displacement which is available through the IO measurement and inter-robot communication. 
	For actual sensors, UWB measurement often has low frame rates, while the frequency of IO is often high. 
	So we model the measurement system as a discrete system.
	On the other hand, when we estimate the real-time relative position of the robot based on Eq. (\ref{Equ_initial}), we can use the IO measurement which is approximately continuous.
	In the next theorem, we will show that the convergence of the proposed estimator (\ref{Equ_observer}) can be guaranteed if the recorded data matrix $S_{ij}$ satisfies the following assumption.
	\begin{assumption}
		\label{A1}
		The rank of the recorded data matrix $S_{ij}$ satisfies $\mathrm{rank}(S_{ij})=2$. 
	\end{assumption}
	
	For the 2-D localization problem, Assumption \ref{A1} indicates that the data matrix $S_{ij}$ records sufficient information to achieve localization.
	This is a common assumption in concurrent-learning methods, which can also be found in \cite{Chowdhary2010CDC} and \cite{Djaneye2019TAC}. The main results regarding the convergence of estimator (\ref{Equ_observer}) are presented in the following theorem.
	
	\begin{theorem}
		\label{Theorem 1}
		Under Assumption \ref{A1}, the estimation error $\tilde{p}_{ij,0}(t)$ is globally exponentially stable with the relative position estimator (\ref{Equ_observer}).
		The convergence rate $\lambda_{ij}$ of $\tilde{p}_{ij,0}(t)$ can be calculated as
		\begin{align}
			\lambda_{ij} \triangleq \sqrt{\left( 1-\frac{\lambda_{\min}(S_{ij})^2}{(2v_{\max}\Delta t + \lambda_{\max}(S_{ij})) ^2} \right)}.
			\label{Equ_converge_rate}
		\end{align}
	\end{theorem}
	\begin{IEEEproof}
		See Appendix \ref{P_T1}.
	\end{IEEEproof}
	
	\begin{remark}
		Theorem \ref{Theorem 1} indicates that when the historical measurement data records sufficient information, i.e. $\mathrm{rank}(S_{ij})=2$, the localization error converges exponentially. Furthermore, the convergence rate of the estimator is highly related to $\lambda_{\min}(S_{ij})$ and $\lambda_{\max}(S_{ij})$. In the next section, we will discuss the relationship between the relative motion of the two robots and the convergence rate of the proposed estimator.
	\end{remark}
	
	\subsection{Localization Enhancement Motion}
	\label{Sec_motion_design}
	In real applications, nevertheless, we still need to address the questions on how to design the relative motion trajectories of two robots and how to collect the measurement data, i.e. how to choose $\varsigma$ and $h$ such that the convergence speed of estimator (\ref{Equ_observer}) is the largest.
	According to Theorem \ref{Theorem 1}, to improve the convergence speed of the estimator, the ratio of the minimum and maximum eigenvalue $\frac{\lambda_{\min}(S_{ij})}{\lambda_{\max}(S_{ij})}$ of the recorded data matrix $S_{ij}$ should be as large as desired. 
	To achieve this, we design the velocity control command of each robot $i$ as the following circular motion velocity 
	\begin{align}
		\label{v_ij}
		v_{i} &= [r_{i}\mathrm{cos}(w_{i}t), r_{i}\mathrm{sin}(w_{i}t)]
	\end{align}
	where $w_{i}=1/i$ is the designed angular velocity and $r_{i}$ is the motion radius. 
	The reason for this design is to enable the estimator (\ref{Equ_observer}) to have a faster convergence speed. 
	For robots $i$ and $j$, we select the measurement interval time $h \Delta t$ and the data set number $\varsigma$ which satisfy
	\begin{subequations}
		\label{con_num_interval}
		\begin{align}
			\mathrm{mod}(2\pi, h \Delta t)&=0 \\
			\varsigma &= ij\varsigma_{0}
		\end{align}
	\end{subequations}
	where $\mathrm{mod}(\cdot)$ is the remainder function and $\varsigma_{0} = \frac{2\pi}{h \Delta t} \in \mathbb{N}$. 
	In the following theorem, we will show that the largest convergence rate of the estimator (\ref{Equ_observer}) can be achieved.
	
	\begin{theorem}
		\label{Theorem 2}
		Under conditions (\ref{v_ij}) and (\ref{con_num_interval}), the proposed estimator (\ref{Equ_observer}) converges with the largest convergence rate, i.e.,  $\frac{\lambda_{\min}(S_{ij})}{\lambda_{\max}(S_{ij})}=1$.
	\end{theorem}
	
	\begin{IEEEproof}
		See Appendix \ref{P_T2}.
	\end{IEEEproof}
	
	\begin{remark}
		Theorem \ref{Theorem 2} shows that with careful design of the relative motion of the two robots, the proposed estimator has the largest convergence rate.
		In the practical applications, the ratio $\frac{\lambda_{\min}(S_{ij})}{\lambda_{\max}(S_{ij})}$ can be calculated at each data measurement time instant $t_{{\rm c},m}$. 
		Define the threshold $\lambda_{0}$ of $\frac{\lambda_{\min}(S_{ij})}{\lambda_{\max}(S_{ij})}$. 	 When $\frac{\lambda_{\min}(S_{ij})}{\lambda_{\max}(S_{ij})}$ is greater than $\lambda_{0}$, the data collection is sufficient, there is no need to maintain the relative motion between the two robots and the estimator will converge exponentially.
		This result directly supports our design of localization enhance command in Section VI B.
	\end{remark}
	
	\section{Behavior-Based Shape Formation}
	\label{Sec_shapeformation} 
	
	Previously, we have shown how to achieve relative localization with only distance and displacement measurements. 
	In this section, we will address the problem of how to achieve shape formation control through behavior based control strategy. 
	To achieve this, we first define and parameterize the desired formation shape. 
	Then, a consensus based shape localization agreement protocol is presented to determine the shape localization in the absence of global coordinates. 
	Thirdly, we propose the distributed shape formation control strategy utilizing local measurements. The details are as follows.
	
	\subsection{Shape Definition and Parameterization}
	
	Firstly, we define the appropriate desired formation.
	\subsubsection{Graphical shape representation}
	
	Similar to our previous work \cite{Sun2023NC}, a user-specified graphical binary image should be provided to represent the desired formation shape. 
	As shown in Fig. \ref{overview}C, the black cells correspond to the desired shape. 
	Each cell is described by two basic parameters $\rho$ and $\xi_{\rho}$. 
	The coordinates $\rho = (\rho_{x}, \rho_{y})$ are the column and row indexes of a cell, see Fig. \ref{overview}C. 
	The scalar $\xi_{\rho} \in [0,1]$ is the color of the cell: $\xi_{\rho} = 0$ if it is black and $\xi_{\rho} = 1$ if it is white. 
	Let $n_{\mathrm{cell}}$ denote the black cell number in the binary image.
	Generally, the larger the value of $n_{\mathrm{cell}}$, the clearer the target image, but it also consumes more storage space.
	Therefore, the desired formation shape can be described by the index set $\mathcal{F} = \{\rho : \xi_{\rho} = 0\}$ and
	$\lvert \mathcal{F} \rvert = n_{\mathrm{cell}}$.
	
	\subsubsection{Gray transformation of the desired formation}
	
	To guide the robot into the desired shape formation smoothly, we convert the binary image to a gray image. Specifically, the set of black cells out by the $l$ cells is expanded to generate an $l$-level gray scale image. 
	For any cell $\rho$ in the image, its gray value can be calculated as,
	\begin{align}
		\label{gray}
		\xi_{\rho}^{k} = \min_{\rho^{'} \in \mathcal{M}_{\rho}} \left ( \xi_{\rho^{'}}^{k-1} + \frac{1}{l} \right ), k=1,2,...,l-1
	\end{align}
	where the superscript $k$ denotes the $k$-th outer layer surrounding the grid, as shown in Fig. \ref{overview}C. The set $\mathcal{M}_{\rho}$ is composed of $3 \times 3$ cells including $\rho$ and its surrounding cells. With iteration (\ref{gray}), the gray formation shape is $\mathcal{F}_{\mathrm{gray}} = \{ \rho: \xi_{\rho}\in[0, 1] \}$.  
	
	\subsubsection{Parameters of the desired formation}
	
	The desired shape represented by image is merely graphical. 
	Each robot should further parameterize it such that it is implementable in physical world.
	The size of the formation shape should be determined.
	The length $l_{\mathrm{cell}}$ of each cell represents the size of the formation shape and can be selected as $l_{\mathrm{cell}} = \sqrt{\frac{\pi}{4}\frac{N}{n_{\mathrm{cell}}}}r_{\mathrm{avoid}}$,
	where $r_{\mathrm{avoid}}$ is the inter-robot collision avoidance distance.
	In this article, we assume that the IO coordinate of each robot has the same direction such that the orientation of the desired shape can be determined as the same.
	Note that it is a common assumption in several relative localization results including \cite{Liu2023Auto} and \cite{Xie2020TRO}, and it can be achieved through different sensors including magnetic compasses.
	
	\subsection{Distributed Shape Localization Agreement}
	\label{Sec_agreement} 
	
	Before introducing the behavior based shape formation strategy, determining the localization of the desired shape formation becomes necessary. However, negotiating the coordinates of the formation localization directly is ineffective due to the lack of global coordinates \cite{Sun2023NC}.
	To handle this issue, we propose a consensus-based shape localization agreement protocol (see Algorithm 1), where the shape localization is marked by the initial position of the seed robot $0$. 
	Before introducing the agreement protocol, we define the concept of neighboring robots first.
	
	\subsubsection{Neighboring robot definition}
	
	The neighbor set $\mathcal{N}_{i}$ of robot $i$ at time instant $t$ is defined as 
	\begin{align}
		\label{neighbor}
		\mathcal{N}_{i} &= \mathcal{N}_{i}^{\mathrm{cur}} \cap \mathcal{N}_{i}^{\mathrm{pas}}
	\end{align}
	with
	\begin{align}
		\notag
		\mathcal{N}_{i}^{\mathrm{cur}} &= \{ j | \lVert p_{i}(t) - p_{j}(t) \rVert \leq r_{\mathrm{sense}}, j \neq i \}
	\end{align}
	and
	\begin{align}
		\notag
		\mathcal{N}_{i}^{\mathrm{pas}} &= \{ j | \exists t'\leq t, \lVert p_{i}(t') - p_{j}(t') \rVert \leq r_{\mathrm{neigh}}, j \neq i \}
	\end{align}
	where $r_{\mathrm{sense}}$ and $r_{\mathrm{neigh}} < r_{\mathrm{sense}}$ denote the maximum measurement radius and a user-defined neighboring radius.
	The set $\mathcal{N}_{i}^{\mathrm{cur}}$ denotes the robots which are in the robot $i$'s sensing range at the current time. 
	The set $\mathcal{N}_{i}^{\mathrm{pas}}$ denotes the robots which are close to robot $i$ in the past and current time. 
	The neighbor set of robot $i$ is defined as the intersection of $\mathcal{N}_{i}^{\mathrm{cur}}$ and $\mathcal{N}_{i}^{\mathrm{pas}}$.
	According to the definition of the neighbor set, it can be seen that when robot $j$ firstly becomes a new neighbor of robot $i$, it is within the circular range of $r_{\mathrm{neigh}}$ of robot $i$ and is unlocalized by robot $i$, i.e., robot $i$ has not collected sufficient measurement data.
	Then they need to collect measurement data to achieve relative localization.
	However, if we directly use whether the distance between robots is less than $r_{\mathrm{sense}}$ as the basis for judging neighbors, the distance $d_{ij}$ between robots $i$ and $j$ may exceed $r_{\mathrm{sense}}$ when robots collect data on relative motion.
	To maintain the measurement for a period of time between robots $i$ and $j$ when they firstly meet each other, we define the set of neighbors as (\ref{neighbor}).
	One can see that communication and measurement between robots $i$ and $j$ can be maintained at least for a time duartion $\frac{r_{\mathrm{sense}} - r_{\mathrm{neigh}}}{2v_{\mathrm{max}}}$.
	
	As shown in Algorithm 1, each robot maintains three sets, i.e., $\mathcal{CD}_{i}$, $\mathcal{RD}_{i}$, $\mathcal{CL}_{i}$ to represent the current measurement data to its neighboring robots, the historical measurement data to neighbors and the localized robot ID. Function $\mathrm{JdgNewNeigh}(\cdot)$ is used to detect the new neighbors which are unlocalized.
	Function $\mathrm{JdgDt}(\cdot)$ judges whether sufficient measurement data is collected, i.e., whether $\frac{\lambda_{\min}(S_{ij})}{\lambda_{\max}(S_{ij})}$ is larger than the data matrix threshold $\lambda_{0}$. 
	If there are new neighbors to be localized or the data collection is insufficient, the localization-enhancing velocity command $v_{i}^{\mathrm{enh}}$ will derive robot $i$ to make localization-enhancing motion within radius $r_{i}$. 
	During this process, robot $i$ collects measurement data (see Algorithm 1, lines 9 - 17).
	The detailed control protocol can be found in Section \ref{control}.
	
	\subsubsection{Finite time consensus based estimator design}
	
	To achieve the shape localization agreement, an estimator $\hat{q}_{i,0}(t)$ is designed to estimate the initial relative position $p_{i0}(t_{0})$ between robot $i$ and the seed robot $0$. The update law $\dot{\hat{q}}_{i,0}(t)$ is designed as
	\begin{align}
		\label{Equ_center}
		\notag
		\dot{\hat{q}}_{i,0}(t)  = &-c_{1} \mathrm{sig} \Big( \sum_{j \in \mathcal{N}_{i}(t_{0})} a_{ij}( \hat{q}_{i,0}(t) - \hat{q}_{j,0}(t) - \hat{p}_{ij,0}(t) ) \\
		& + \mu_{i} ( \hat{q}_{i,0}(t) - \hat{p}_{i0,0}(t) ) \Big)^{\alpha}
	\end{align}
	where $c_{1}>0$ and $0<\alpha<1$ are the gain and exponent of the estimator (\ref{Equ_center}). 
	Parameter $a_{ij} = 1$ means that robot $j$ is a neighbor of robot $i$ at initial time $t_{0}$, otherwise $a_{ij} = 0$.
	Parameter $\mu_{i}=1$ means that robot $i$ is a neighbor of the seed robot, otherwise $\mu_{i} = 0$. Define $\mathcal{B} = \mathrm{diag}(\mu_{1},...,\mu_{n})$. 
	For vector $x=[x_{1},...,x_{n}]^T$, function $\mathrm{sig}(x)^{\alpha} = \sum_{i=1}^{n}\mathrm{sign}(x_{i})\lvert x_{i} \rvert^{\alpha}$, where $\mathrm{sign}(\cdot)$ is the sign function. 
	Recall that the relative position estimator $\hat{p}_{ij,0}$ defined in (\ref{Equ_observer}) is also adopted in (\ref{Equ_center}).
	Define the estimated error as $\tilde{q}_{i,0} = \hat{q}_{i,0} - p_{i0}(t_{0})$, take $\tilde{q} = [\tilde{q}_{1,0},...,\tilde{q}_{n,0}]^T$.

	Based on the initial state estimator (\ref{Equ_center}), the real-time shape localization estimate is designed as,
	\begin{align}
		\label{Equ_location2}
		\hat{q}_{i}(t) = \hat{q}_{i,0}(t) + z_{i}(t).
	\end{align} 
	Recall that $z_{i}(t)$ is the displacement measured through IO of robot $i$. The real-time estimator designed as (\ref{Equ_location2}) is due to the fact that the initial position of the seed robot, i.e., shape localization is static. Note that the initial neighbor set $\mathcal{N}_{i}(t_{0})$ is adopted in (\ref{Equ_center}). Before introducing the main result, the following assumption should be satisfied.
	\begin{assumption}
		\label{A2}
		The initial measurement topology $\mathcal{G}(t_{0})$ defined by (\ref{neighbor}) is connected and the neighboring radius $r_{\mathrm{neigh}}$ satisfies $r_{\mathrm{neigh}} < r_{\mathrm{sense}} - r_{i} - r_{j}$, for $j \in \mathcal{N}_{i}(t_{0})$.
	\end{assumption}
	
	Note that the connected constrain of $\mathcal{G}(t_{0})$ in Assumption \ref{A2} is common and can be found in several consensus results including \cite{Liu2023Auto,Xie2020TRO}. 
	The constrain of $r_{\mathrm{neigh}}$ is to guarantee that the initial neighbor robots in set $\mathcal{N}_{i}(t_{0})$ are still the neighbors of robot $i$ during the shape agreement process.
	Let $\mathcal{L}$ be the Laplace matrix of the initial measurement topology $\mathcal{G}(t_{0})$.
	The convergence of the agreement protocol (\ref{Equ_center}) can be given in the following theorem. 
	Choose a Lyapunov function candidate,
	\begin{align}
		\label{V_l}
		V_{l} = \frac{1}{2} \tilde{q}^T (\mathcal{L} + \mathcal{B}) \tilde{q}
	\end{align} 
	
	\begin{algorithm}[!t]
		\small
		\caption{Shape localization agreement}
		\KwIn{data matrix threshold $\lambda_{0}$, threshold convergence rate $\delta_{0}$, swarm scale $n$.}
		\KwOut{shape relative localization estimator $\hat{q}_{i,0}$}
		\BlankLine    
		$\mathcal{N}_{i}(t_{0})$ $\leftarrow$ get robot $i$'s neighbor set \;
		$hop_{i}$ $\leftarrow$ set as $0$ \;
		$\mathcal{CD}_{i}$, $\mathcal{RD}_{i}$, $\mathcal{RL}_{i}$ $\leftarrow$ set as empty set $\phi$ \;
		\While{ture}{
			\If{$hop_{i}<n$}{
				\scriptsize{\tcp*[h]{Check whether the agreement is completed var the hop-count}\;}
				$\mathcal{N}_{i}$ $\leftarrow$ get robot $i$'s neighbor set \;
				\eIf{$\mathrm{JdgNewNeigh}(\mathcal{N}_{i}, \mathcal{RL}_{i})$ is false and $\mathrm{JdgDt}$($i$, $\mathcal{RD}_{i}$, $\lambda_{0}$) is ture}{
					\scriptsize{\tcp*[h]{No neighbor to be localized and current neighbors have been localized}\;}
					$\mathcal{RL}_{i}$ $\leftarrow$ update the localized robot set \;
					$v_{i}$ $\leftarrow$ set as $0$\;
				}{
					\scriptsize{\tcp*[h]{Robot $i$ detects unlocalized neighbors, localization enhance and collect data}\;}
					$\mathcal{CD}_{i}$ $\leftarrow$ get current distance and displacement measurements of robot $i$ and its neighbors\;
					$\mathcal{RD}_{i}$ $\leftarrow$
					record current mearsurement $\mathcal{CD}_{i}$\;
					$v_{i}^{\mathrm{enh}}$ $\leftarrow$ calculate localization enchance command (\ref{v_enh})\;
					$v_{i}$ $\leftarrow$ $v_{i}^{\mathrm{enh}}$\;
				}
				\If{$\mathrm{CalUpdRate}$($\hat{q}_{i,0}$) $> \delta_{0}$}{
					$hop_{i}$ $\leftarrow$ set as $0$
				}
				\Else{
					$hop_{i}$ $\leftarrow$  $\mathop{\min}_{j \in \mathcal{N}_{i}}(hop_{j}) + 1$
				}
				\scriptsize{\tcp*[h]{Relative localization and shape position agreement}\;}
				\For{$j \in \mathcal{RL}_{i}$}{
					$\hat{p}_{ij,0}$ $\leftarrow$ update relative position estimator according to (\ref{Equ_observer})
				}
				$\hat{q}_{i,0}$ $\leftarrow$ update shape position estimator according to (\ref{Equ_center})\;
			}
			\Else{
				return $\hat{q}_{i,0}$
			}
		}
	\end{algorithm}
	
	\begin{theorem}
		\label{theorem 3}
		Under Assumption \ref{A2}, given $(i,j) \in \mathcal{G}(t_{0})$, the measurement data matrix $S_{ij}$ satisfies Assumption \ref{A1}.
		As a result, the agreement error $\tilde{q}$ will converge into an adjustable set $\Omega = \{ \lVert \tilde{q} \rVert \leq b \}$ in a finite time $t_{a} + t_{l}$ with the estimation law (\ref{Equ_center}), where $b$, $t_{a}$ and $t_{l}$ are defined as,
		\begin{align}
			\label{b}
			b = \sqrt{\frac{ 2^{\frac{1+3\alpha}{2\alpha(\alpha+1)}} \left( \sum_{i=1}^{n} (|\mathcal{N}_{i}(t_{0})|\varepsilon)^{1+\alpha} \right)}{(1-\gamma) \lambda_{\min}(\mathcal{L} + \mathcal{B})^2}}
		\end{align}
		\begin{align}
			\label{t_l}
			t_{l} = \frac{2(1+\alpha)V_{l}(t_{a})^{\frac{1-\alpha}{2}}}{\lambda_{\min}(\mathcal{L} + \mathcal{B})\gamma(1-\alpha)}
		\end{align}
		\begin{align}
			\label{t_a}
			t_{a} = \mathrm{\max}\{ \frac{\mathrm{ln} (\varepsilon \lVert \tilde{p}_{ij,0}(t_{0}) \rVert^{-1})}{\mathrm{ln}(\lambda_{ij})} \Delta t, (i,j) \in \mathcal{G}(t_{0})\}
		\end{align}
		where $|\mathcal{N}_{i}(t_{0})|$ represents the initial neighbor number of robot $i$. 
		Positive constants $\varepsilon>0$ and $\gamma>0$ are adjustable parameters. The convergence rate $\lambda_{ij}$ is defined in (\ref{Equ_converge_rate}).
	\end{theorem}
	
	\begin{IEEEproof}
		See Appendix \ref{P_T3}.
	\end{IEEEproof}
	
	\begin{remark}
		Note that $V_{l}(t_{a})$ in (\ref{t_l}) can be calculated with the initial sate $V_{l}(t_{0})$ and initial relative localization error $\tilde{p}_{ij,0}(t_{0})$ according to (\ref{V_l}). 
		Theorem \ref{theorem 3} offers an answer to the question how long it takes to ensure that the agreement error $\tilde{q}$ is less than an adjustable constant $b$ defined in (\ref{b}). In other words, Theorem \ref{theorem 3} shows how long it takes for the shape localization achieves an agreement. 
	\end{remark} 
	
	However, calculating convergence time requires the initial measurement topology of the swarm, which is unavailable in real applications.  
	To overcome this issue, a gradient threshold based decision method and the hop-count information transmission algorithm \cite{Wang2020TRO} are adopted in this article. 
	Function $\mathrm{CalUpdRate(\cdot)}$ calculates the variation rate of $\hat{q}_{i,0}$ over a period of time $\delta_t$ which can be defined as,
	\begin{align}
		\label{variation_rate}
		\Delta \hat{q}_{i,0} = \frac{\lVert \hat{q}_{i,0}(t) - \hat{q}_{i,0}(t - \delta_t) \rVert}{\delta_t}.
	\end{align}
	
	If $\Delta \hat{q}_{i,0}$ defined as (\ref{variation_rate}) remains above a user-defined threshold convergence rate $\delta_{0}$, the hop-count $hop_{i}$ is set as 0. Otherwise, it means that the robots have reached an approximate agreement, the hop-count $hop_{i}$ is set as $\mathrm{\min}(hop_{j})+1$, where $j \in \mathcal{N}_{i}(t_{0})$ (Algorithm 1, Lines 18-22). 
	The value of hop-count increases rapidly when all robots reach an agreement.
	
	\subsection{Distributed Shape Formation}
	\label{control}
	
	The proposed behavior based control strategy consists of two stages, i.e., the shape localization agreement (Algorithm 1, lines 9-17) and shape formation (Algorithm 2). 
	During the shape localization agreement stage, if robot $i$ does not have unlocalized neighbors,
	the control law $v_{i}$ is designed as
	\begin{align}
		\label{control_s1_1}
		v_{i} = 0.
	\end{align}
	Otherwise, robot $i$ makes circular motion to enhance localization with neighboring robots, i.e., $v_{i}$ is designed as
	\begin{align}
		\label{control_s1_2}
		v_{i} = v_{i}^{\mathrm{enh}}
	\end{align}
	where $v_{i}^{\mathrm{enh}}$ is the localization-enhancing command.
	
	When all the robots reach a consensus on the shape localization with the agreement protocol (\ref{Equ_center}).
	The shape formation stage begins. 
	During this stage, if robot $i$ does not have unlocalized neighboring robots, $v_{i}$ is designed as
	\begin{align}
		\label{control_s3_1}
		v_{i} = v_{i}^{\mathrm{ent}} + v_{i}^{\mathrm{exp}} + v_{i}^{\mathrm{int}}.
	\end{align}
	Otherwise, the control law is designed as
	\begin{align}
		\label{control_s3_2}
		v_{i} = v_{i}^{\mathrm{int}} +  v_{i}^{\mathrm{enh}}.
	\end{align}
	Here, $v_{i}^{\mathrm{ent}}$, $v_{i}^{\mathrm{exp}}$ and $v_{i}^{\mathrm{int}}$ denote the shape-entering command, shape-exploring command and neighbor-interacting command.
	
		\begin{algorithm}[!t]
		\small
		\caption{Distributed shape formation control}
		\KwIn{shape relative localization estimator $\hat{q}_{i,0}$,
			data matrix threshold $\lambda_{0}$,
			localized robot set and collected measurement data $\mathcal{RL}_{i}$, $\mathcal{RD}_{i}$.}
		\BlankLine    
		
		$\mathcal{V}_{i}$ $\leftarrow$ set as empty set $\phi$ \;
		\While{ture}{
			$\mathcal{N}_{i}$ $\leftarrow$ get robot $i$'s neighbor set \;
			
			\If{$\mathrm{JdgNewNeigh}(\mathcal{N}_{i})$ is false and $\mathrm{JdgDt}$($i$, $\mathcal{RD}_{i}$,$\lambda_{0}$) is ture}{
				\scriptsize{\tcp*[h]{No neighbor to be localized and current neighbors have been localized}\;}
				$\mathcal{RL}_{i}$ $\leftarrow$ update the localized robot set\;
				\scriptsize{\tcp*[h]{Behavior based shape formation control}\;}
				$v_{i}^{\mathrm{ent}}$ $\leftarrow$ calculate shape forming command (\ref{v_ent})\;
				$v_{i}^{\mathrm{exp}}$ $\leftarrow$ calculate shape exploration command (\ref{v_exp})\;
				$v_{i}^{\mathrm{int}}$ $\leftarrow$ calculate interaction command (\ref{v_int})\;
				$v_{i}$ $\leftarrow$ $v_{i}^{\mathrm{ent}}$ + $v_{i}^{\mathrm{exp}}$ + $v_{i}^{\mathrm{int}}$;
			}
			\Else{
				\scriptsize{\tcp*[h]{Robot $i$ detects unlocalized neighbors, localization enhance and collect data}\;}
				
				$\mathcal{CD}_{i}$ $\leftarrow$ get current distance and displacement measurements of robot $i$ and its neighbors\;
				$\mathcal{RD}_{i}$ $\leftarrow$
				record current measurement $\mathcal{CD}_{i}$\;
				$v_{i}^{\mathrm{enh}}$ $\leftarrow$ calculate localization enhance command (\ref{v_enh})\;
				$v_{i}^{\mathrm{int}}$ $\leftarrow$ calculate interaction command (\ref{v_int})\;
				$v_{i}$ $\leftarrow$ $v_{i}^{\mathrm{enh}} + v_{i}^{\mathrm{int}}$ \;
			}
			
			$\mathcal{V}_{i}$ $\leftarrow$ get velocity command of robot $i$'s neighbors.
			\scriptsize{\tcp*[h]{Relative localization during shape formation}\;}
			\For{$j \in \mathcal{RL}_{i}$}{
				$\hat{p}_{ij,0}$ $\leftarrow$ update relative position esetimator according to (\ref{Equ_observer}) 
			}
		}
	\end{algorithm}
	
	According to the control law designed as (\ref{control_s1_1}) and (\ref{control_s1_2}), one can see that each robot either performs circular localization-enhancing motion or stops during the shape localization agreement stage. 
	As a result, the collision avoidance between robots can be achieved if the initial distance $d_{ij}(t_{0})$ between them satisfies $d_{ij}(t_{0}) > 2(r_{i} + r_{j})$.  
	During the shape formation stage, the collision avoidance between robots can be achieved by the neighbor-interacting command $v_{i}^{\mathrm{int}}$ in (\ref{control_s3_1}) and (\ref{control_s3_2}).
	The four control commands are designed as follows.
	
	\subsubsection{Shape-entering command}
	When all the neighboring robots of robot $i$ are localized, robot $i$ turns to form the desired shape (Algorithm 2, lines 4-11). 
	The shape-entering command $v_{i}^{\mathrm{ent}}$ is designed to derive robot $i$ to its interpretation of the desired shape, as depicted in Fig. \ref{overview}C.
	The shape-entering command can also be understood as a proportional controller, driving the robots into the shape.
	
	It is designed as
	\begin{align}
		\label{v_ent}
		v_{i}^{\mathrm{ent}} = \kappa_{1} \xi_{\rho_{T,i}} \frac{\hat{q}_{i} + p_{\rho_{T,i}0}}{\hat{q}_{i} + p_{\rho_{T,i}0}}
	\end{align}
	where $\kappa_{1}>0$ is the gain of the shape-entering command.The larger $\kappa_{1}$, the greater the force that pushes the robot into shape.
	Recall that $\hat{q}_{i}$ is the real-time shape localization estimate (\ref{Equ_location2}).
	The relative position between the nearest gray cell to robot $i$ is denoted as $p_{\rho_{T,i}0}$.
	Note that $p_{\rho_{T,i}0}$ can be determined by $\hat{q}_{i}$ and the gray formation shape $\mathcal{F}_{\mathrm{gray}}$.
	The gray level of the cell is $\xi_{\rho_{T,i}}$.
	\subsubsection{Shape-exploring command}
	
	The shape-exploration command $v_{i}^{\mathrm{exp}}$ aims to guide robot $i$ into the desired formation shape and explore the unoccupied black region, see Fig. \ref{overview}C. Note that the mean-shift concept is adopted to design $v_{i}^{\mathrm{exp}}$ as,
	\begin{align}
		\label{v_exp}
		v_{i}^{\mathrm{exp}} = \frac{\sum_{\rho \in \mathcal{M}_{i}^{\mathrm{sense}}} \kappa_{2} \phi (\lVert \hat{q}_{i} + p_{\rho 0} \rVert) \left( \hat{q}_{i} + p_{\rho 0} \right) }{\sum_{\rho \in \mathcal{M}_{i}^{\mathrm{sense}}} \phi (\lVert \hat{q}_{i} + p_{\rho 0} \rVert) }
	\end{align}
	where $\kappa_{1}>0$ is the gain of the shape-exploring command. The larger $\kappa_{2}$, the greater the force exerted by the robot in shape exploration. Function $\phi(z)$ is defined as,
	\begin{align}
		\begin{aligned}
			\phi(z) = \left \{
			\begin{aligned}
				\notag
				&1 &z \leq 0 \\
				&\frac{1}{2}(1+\mbox{cos}\pi z) \quad &0<z<1 \\
				&0 &z \geq 1
			\end{aligned}
			\right. 
		\end{aligned}
	\end{align}
	where $\mathcal{M}_{i}^{\mathrm{sense}}$ is defined as the set of unoccupied black cells that are in the sensing radius $r_{\mathrm{sense}}$ of robot $i$. Note that $\mathcal{M}_{i}^{\mathrm{sense}}$ can be determined through $\hat{q}_{i}$ and $\hat{p}_{ij}$, for $j\in \mathcal{N}_{i} \cap \mathcal{RL}_{i}$. 
	Let $\rho \in \mathcal{M}_{i}^{\mathrm{sense}}$ denote the unoccupied black cells and $p_{\rho 0}$ represent the relative position between the $\rho$ cell and shape localization. Similarly, $p_{\rho 0}$ can be determined by the gray formation shape $\mathcal{F}_{\mathrm{gray}}$.
	
	\subsubsection{Neighbor-interacting command}
	As shown in Fig. \ref{overview}C, the interaction command $v_{i}^{\mathrm{int}}$ is to achieve collision avoidance and velocity alignment between robots, it is designed as 
	\begin{align}
		\label{v_int}
		v_{i}^{\mathrm{int}} = \kappa_{3} \sum_{j \in \mathcal{N}_{i} \cap \mathcal{RL}_{i}} \mu(d_{ij})\hat{p}_{ij} + \kappa_{4} \sum_{j \in \mathcal{N}_{i} \cap \mathcal{RL}_{i}} (v_{i} - v_{j})
	\end{align}
	where $\kappa_{3}>0$ and $\kappa_{4}>0$ are the gains of collision-avoidance and velocity-alignment, respectively. The larger $\kappa_{3}$, the greater the collision avoidance force between robots, similar to $\kappa_{4}$. $\mu(z)$ is defined as
	\begin{align}
		\mu(z) = \left \{
		\begin{aligned}
			\notag
			&\frac{r_{\mathrm{avoid}}}{z} - 1 &z \leq r_{\mathrm{avoid}} \\
			&0  & z>r_{\mathrm{avoid}}
		\end{aligned}.
		\right. 
	\end{align}
	
	For robot $i$, when calculating the interaction force, only the localized neighboring robots in set $\mathcal{RL}_{i}$ are considered. 
	As for the current neighboring robots which are not in the localized neighbor set $\mathcal{RL}_{i}$, they are often far away from robot $i$ and the collision risk is low.
	
	\subsubsection{Localization-enhancing command}
	During the process of shape formation control, when robot $i$ encounters a neighboring robot $j$ which is not localized, i.e., $j \in \mathcal{N}_{i}$ and $j \notin \mathcal{RL}_{i}$. 
	$v_{i}^{\mathrm{enh}}$ drive it to make localization-enhancing motion and collect measurement data to achieve localization(Algorithm 2, Lines 12-18).
	According to the theoretical analysis in Section \ref{Sec_motion_design}, localization-enhancing command $v_{i}^{\mathrm{enh}}$ is designed as
	\begin{align}
		\label{v_enh}
		v_{i}^{\mathrm{enh}} &= [r^{'}_{i}w_{i}\mathrm{cos}(\varphi), r^{'}_{i}w_{i}\mathrm{sin}(\varphi)]^T
	\end{align}
	with
	\begin{align}
		\notag
		r^{'}_{i} &= \left\{
		\begin{aligned}
			&r_{i}, &r_{i} \leq \mathrm{\min}\{\mathcal{D}_{i}\} \\
			&\mathrm{\min}\{\mathcal{D}_{i}\}, &\mathrm{Otherwise}
		\end{aligned}
		\right. 
	\end{align}
	and 
	\begin{align}
		w_{i} &= w_{r} \mathrm{rand}(0,1) + w_{0} / i
	\end{align}
	where $\varphi$ is the heading angle of robot $i$. The desired radius of the circular motion and the real radius of the circular motion are defined as $r_{i}$ and $r^{'}_{i}$, respectively. Data set $\mathcal{D}_{i} \subseteq \mathcal{CD}_{i}$ denotes the neighboring robots' distance measurement. Function $\mathrm{rand}(0,1)$ returns a random value from 0 to 1 and $w_{r}>0$ is a constant. The random term $w_{i}$ is designed to accelerate the data collection of the robots whose ID value is large.
	
	\begin{figure}[!t]\centering
		\includegraphics[scale=0.97]{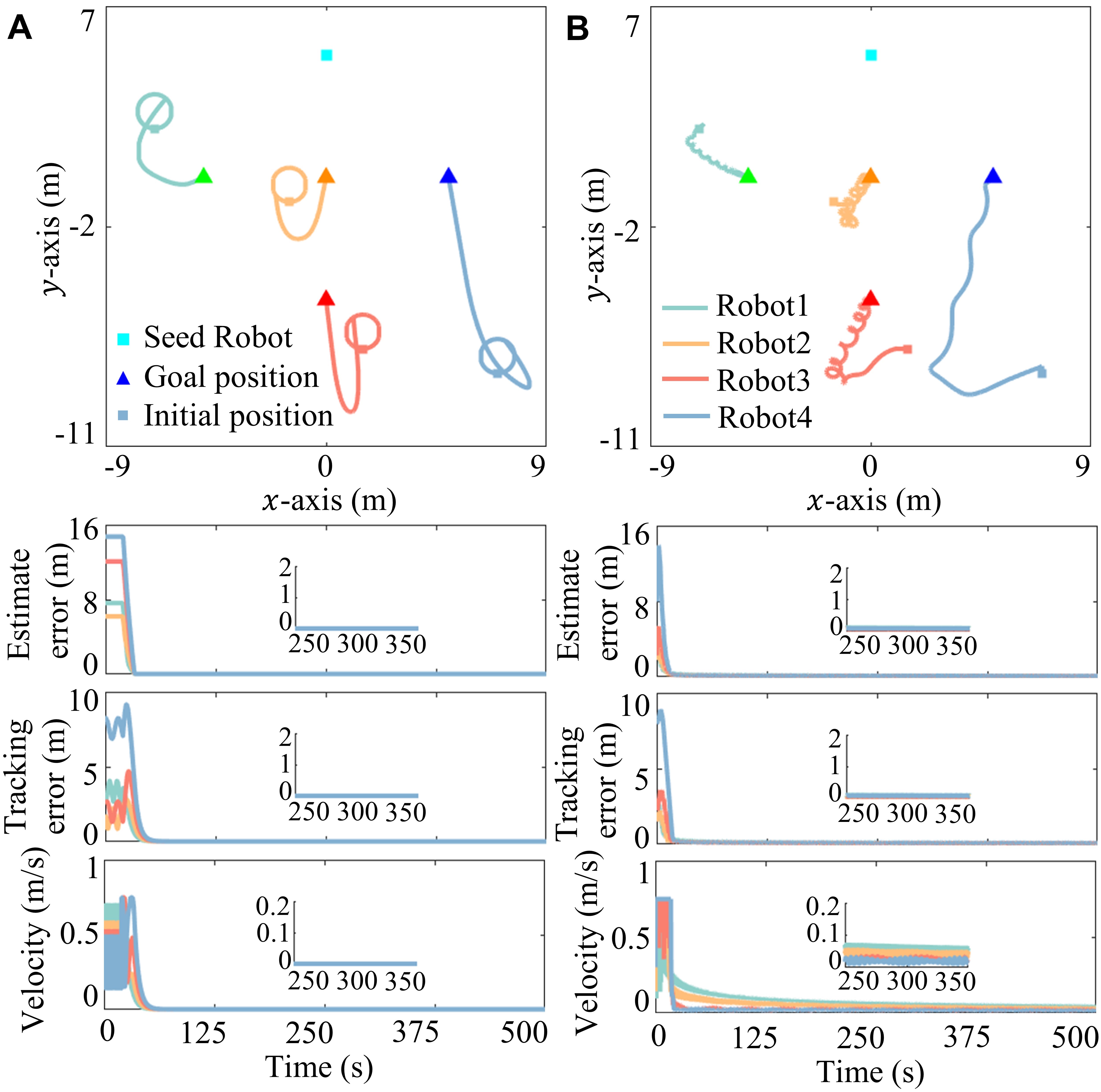}
		\caption{Comparision between the proposed relative localization method and the PE based one in \cite{Xie2019TCNS}. 
		\textbf{A}. Simulation results with the proposed method, from top to bottom are the trajectory, estimate error, tracking error and velocity norm of each robot. Here, tracking error and estimation error are defined as $e_{i}^{\mathrm{c}}(t) = \lVert {p}_{i}(t) - p_{0} - p_{i} \rVert$ and $e_{i}^{\mathrm{e}}(t) = \lVert \hat{p}_{i,0}(t) - p_{0} - p_{i} \rVert$, respectively.
		\textbf{B}. Simulation results with method in \cite{Xie2019TCNS}.
		}\centering
		\label{S1_1}
		\vspace{-0.5cm}
	\end{figure}
	
	The localization-enhancing command derives robot $i$ to make circular motion, as shown in Fig. \ref{overview}C. According to Theorem \ref{Theorem 2}, this command helps to improve the convergence speed of the proposed localization algorithm.
	
	Compared with our previous work \cite{Sun2023NC}, we
	focus on the shape formation control with only local measurements in this article. 
	To achieve this goal, a localization-enhancing distributed control strategy is proposed. All the control commands are based on the estimated relative position. 
	The relative position estimator (\ref{Equ_observer}) with exponential convergence and the center negotiation estimator (\ref{Equ_center}) with finite time convergence play an important role in the close-loop control strategy.
	
	\section{Simulation Results and Performance Analysis}
	\label{Sec_simulation}
	Simulation results are given in this section to show the robustness, effectiveness and adaptability of our methods.    
	
	\begin{figure}[!t]\centering
		\includegraphics[scale=0.97]{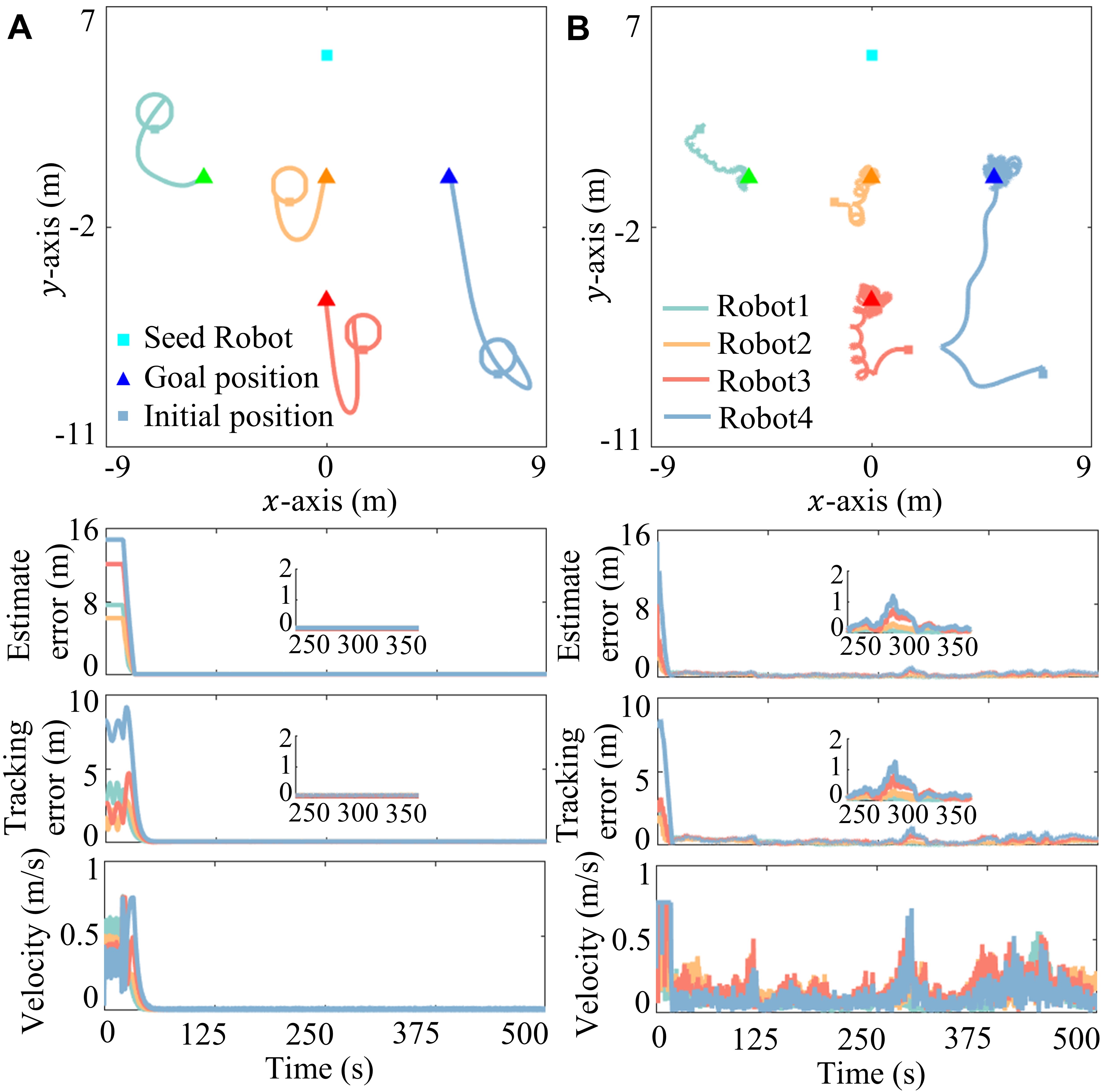}
		\caption{Comparision between the proposed relative localization method and the PE based one in \cite{Xie2019TCNS} with the sensor measurement noise. 
		\textbf{A}. 
		Simulation results with the proposed method, from top to bottom are the trajectory, estimate error, tracking error and velocity norm of each robot. 
		\textbf{B}. 
		Simulation results with method in \cite{Xie2019TCNS}.
		}\centering
		\label{S1_2}
		\vspace{-0.5cm}
	\end{figure}
	
	\subsection{Performance of Relative Localization Method}
	
	First, we evaluate the performance of the proposed relative localization method by a distributed formation scenario.
	In the case without measurement noise, five robots are assigned to achieve formation with the proposed concurrent-learning based relative localization method and the localization method proposed in \cite{Xie2019TCNS}. 
	The position of a seed robot is chosen as $ p_{0} = [0,7]^T$ and the reference formation offsets for another four robots are chosen as $p* = [p_{1}^{*T}, p_{2}^{*T}, p_{3}^{*T}, p_{4}^{*T}]^T=[-7,-7,0,-7,0,-14,7,-7]^T$. 
	The measurement topology is fixed, the connectivity matrix $\mathcal{A} $ and informed matrix $\mathcal{B}$ are $\mathcal{A}=[0, 1, 0, 0; 1, 0, 1, 0; 0, 1, 0, 1; 0, 0, 1, 0]$ and $\mathcal{B} = \mathrm{diag}(1,1,0,0)$. 
	The control input of each robot is $v_{i}(t) = -\kappa(\hat{q}_{i}(t) - p_{i}^{*})$,
	where $\kappa=0.2$.
	The localization enhancement motion radius and angle velocity are $r_{i} = 0.6$m, $v_{\mathrm{max}}=0.75$m/s, $w_{0}=1$rad/s, $w_{r}=1$rad/s. 
	The parameters adopted in the relative position estimator are $\Delta t=0.01$ $\lambda_{0}=0.1$, $c_{1} = 0.005$, $\alpha=0.1$, and the data batch size is chosen as $h=60$.
	Note that the controller here is not the focus of our simulation. 
	We have chosen the simplest P-controller, and other controllers such as PID or MPC can also be used in practical use to improve tracking control performance.
	Under the fixed communication and measurement topology, there is no need for parameters $\delta_t$ and $\delta_{0}$ in this scenario.
	The parameters adopted in the comparative simulation can be found in \cite{Xie2019TCNS}. 
	The trajectories of the robots and the tracking errors are shown in Fig. \ref{S1_1}A and \ref{S1_1}B. It can be seen that without the persistent noise, the trajectories of the robots become smoother and velocity command converges to zero quickly in Fig. \ref{S1_1}A.

	Furthermore, to verify the noise robustness of the relative localization algorithm, white Gaussian noises with zero mean and variance of $0.02$m and $0.002$m/s are added to distance and odometry measurement, respectively. 
	The parameters adopted in this case are the same as those in the noise-free case. 
	The trajectories of the robots, the norm of tracking errors, the norm of estimation errors and the norm of robot velocities are shown in Fig. \ref{S1_2}C and \ref{S1_2}D.
	From the simulation results, it can be seen that compared with the method proposed in \cite{Xie2019TCNS}, the concurrent-learning based estimation algorithm has better robustness against noisy measurements. 
	This is due to the fact that our method simultaneously utilizes historical data for information enhancement.
	
	\subsection{Effectiveness of the Shape Formation Scheme}
	\label{sec_S2}
	\begin{figure}[!t]\centering
		\includegraphics[scale=1.055]{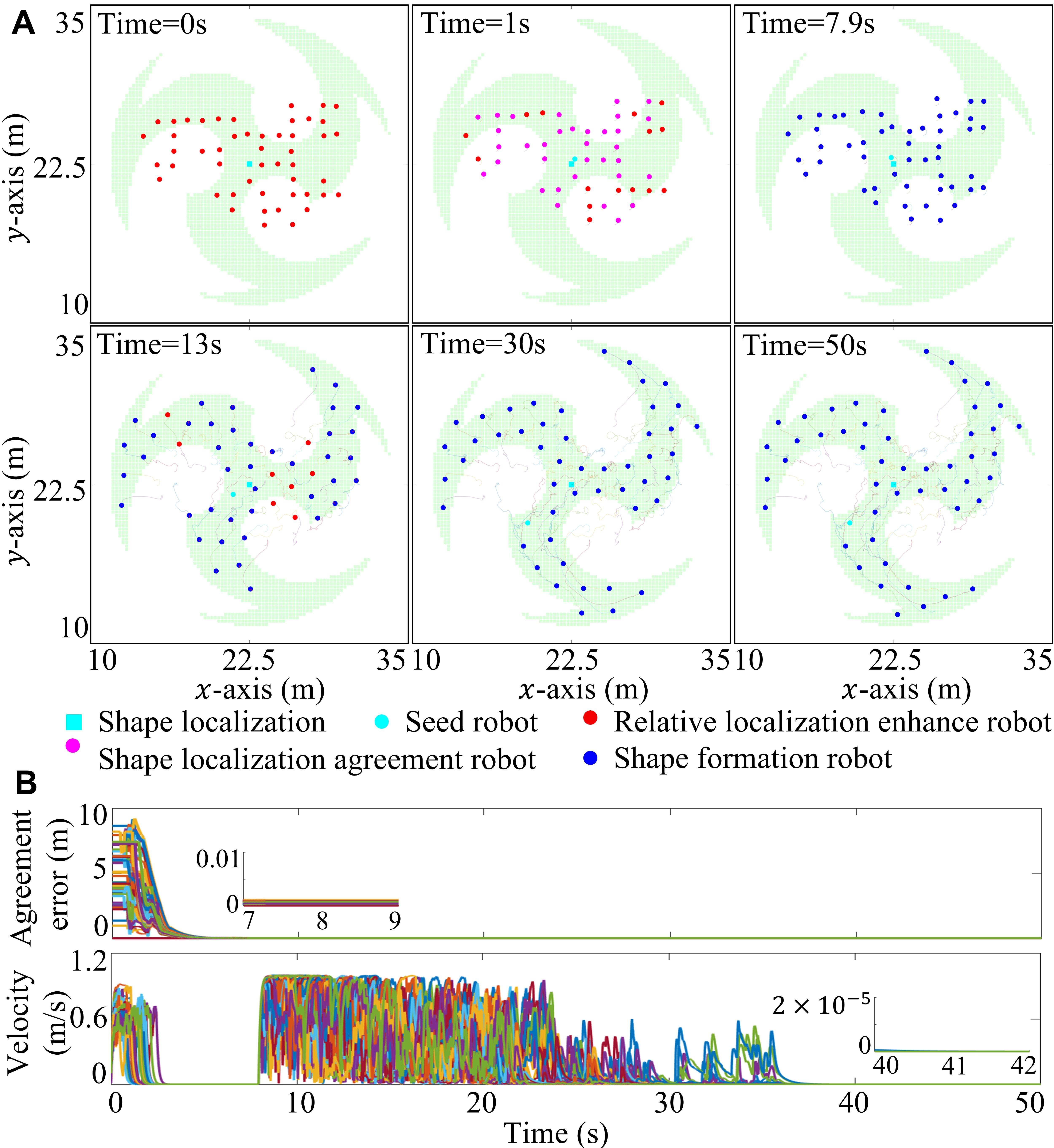}
		\caption{Shape formation process for a swarm of 50 robots. \textbf{A}. Snapshots of the shape formation process. From left to right and from top to bottom are the representative moments of the robots' states, positions, and motion trajectories.
		\textbf{B}. Shape localization agreement error and velocity command of each robot.}\centering
		\label{S2}
		\vspace{-0.5cm}
	\end{figure}

	In the swarm shape formation scenario, a swarm of 50 robots is assigned to form a dart shape. 
	Compared with the latest result \cite{Wang2020TRO} and our previous work \cite{Sun2023NC}, only onboard UWB and IO measurements are used to achieve the shape formation task for robot swarms.
	The parameters $r_{\mathrm{sense}}$ and $r_{\mathrm{neigh}}$ are chosen as $r_{\mathrm{sense}}=4$m and $r_{\mathrm{neigh}}=2.5$m. 
	Furthermore, the collision avoidance radius is set as $r_{\mathrm{avoid}}=1.8$m. 
	The localization enhancement motion radius and angle velocity are $r_{i} = 0.3$m, $v_{\mathrm{max}}=1$m/s, $w_{0}=6$rad/s, $w_{r}=4$rad/s.
	The parameters adopted in the relative localization are $\Delta t=0.05$s, $h=40$, $\lambda_{0}=0.1$, $c_{1}=0.1$, $\alpha=0.5$, $\delta_t = 1$s, $\delta_{0}=0.01$m. 
	The parameters used in the behavior control are $n_{\mathrm{cell}}=2130$, $\kappa_{1}=10$, $\kappa_{2}=15$, $\kappa_{3}=25$, $\kappa_{4}=2$.
	For more information about parameter selection, please refer to Appendix \ref{parameter_discussion}.

	The process of shape formation is shown in Fig. \ref{S2}A. 
	During the shape forming process, the different states of each robot are represented by three colors. 
	Red represents that the robot is in neighbor relative localization state and making relative motion to collect measurement data. According to Algorithms 1 and 2, one can see that the velocity command of the red robot is $v_{i} = v_{i}^{\mathrm{enh}} + v_{i}^{\mathrm{int}}$ or $v_{i} = v_{i}^{\mathrm{enh}}$. 
	Magenta represents that the robot is in shape localization agreement state. 
	In this state, according to Algorithm 1, the velocity command is $v_{i}=0$. 
	Blue represents that the robot is in distributed shape formation state. 
	In this state, according to Algorithm 2, the velocity command is $v_{i}=v_{i}^{\mathrm{form}} + v_{i}^{\mathrm{exp}} + v_{i}^{\mathrm{int}}$.
	
	At time instant $t=0$s, each robot moves and collects measurement data to estimate the relative position to its neighbor robots. 
	As some robots complete the relative localization of their neighboring robots at $t=1$s, the process of the shape localization agreement begins.
	Although some robots are still in red, they will turn to magenta when they collect sufficient data. 
	After the convergence of the seed robot relative position estimator $\hat{q}_{i}(t)$ for each robot $i$, the robots turn into distributed shape formation control state under the action of the hop-count algorithm at $t=7.6$s. 
	The norm value of the shape localization agreement errors $\tilde{q}_{i,0}$ is shown in Fig. \ref{S2}B. 
	One can see that with the proposed shape localization agreement protocol and the hop-count based end-state decision method, the agreement errors are less than 0.5cm. 
	During the process of the shape formation, some robots may encounter new neighbors and re-enter the neighbor relative localization state, as shown in the snapshot of the robots' states at $t=13$s. 
	It can be seen that the desired shape is almost formed by the robot swarms at $t=30$s.
	According to the norm value of velocity $v_{i}$ shown in Fig. \ref{S2}B, the robot swarms are static at $t=40$s, which means the shape formation process is completed. 
	With the proposed methods, the robot swarms form the desired dart shape at $t=50$s. 
	
	\subsection{Adaptability to Swarm Scale Variants}
	
\begin{figure}\centering
		\includegraphics[scale=1.045]{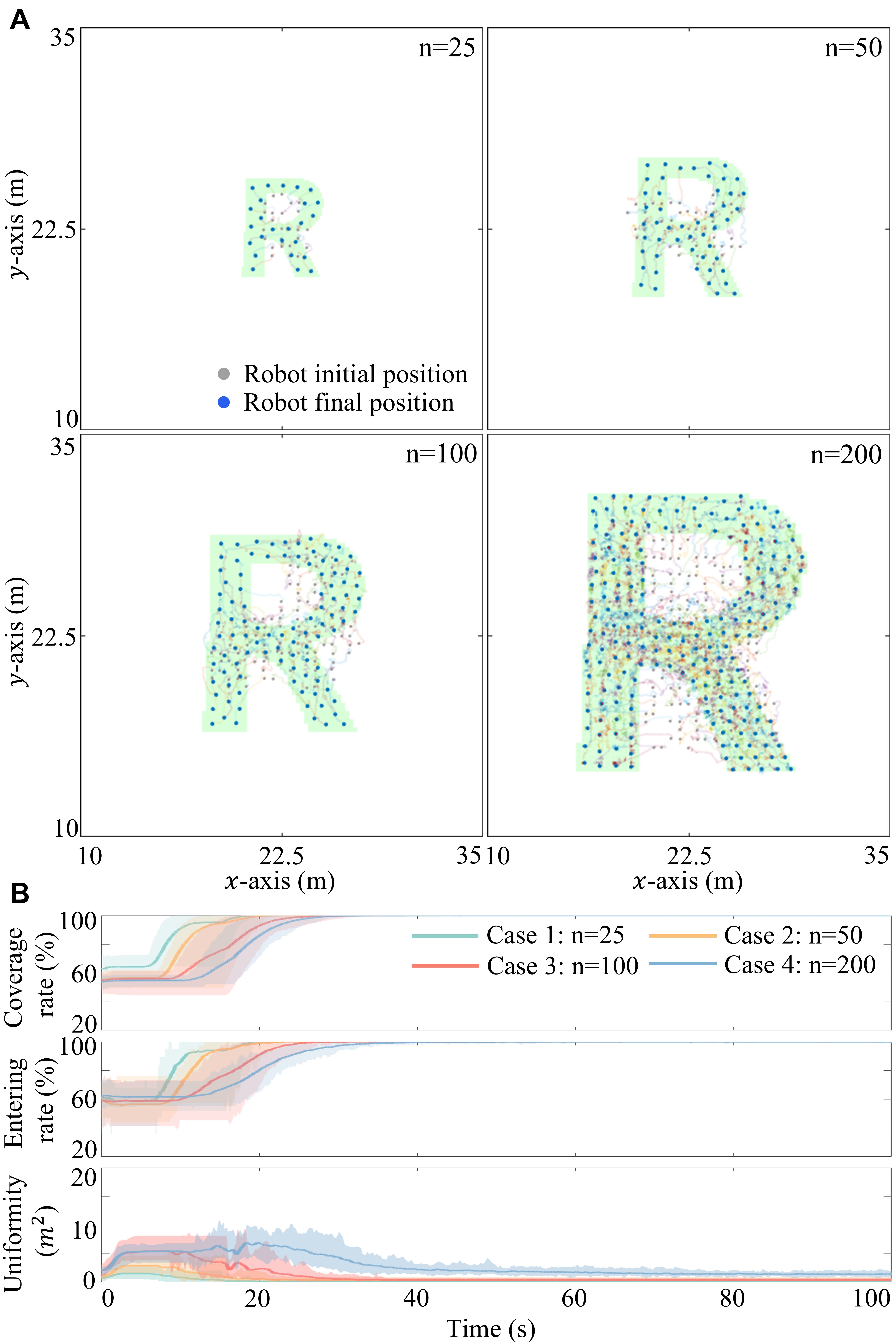}
		\caption{Simulation results to evaluate the adaptability of the proposed method given different swarm scale $n$. 
				A. Trajectory and final position of the swarm with 25, 50, 100 and 200 robots.
				B. Coverage rate, entering rate and distribution uniformity of the robot swarms to form capital letter ``R''.
		}\centering
		\label{S3}
		\vspace{-0.5cm}
	\end{figure}

	In the adaptability test scenario, robot swarms with different scales are assigned to form the capital letter ``R'' shape.
	For the robot swarms with scale 25, 50, 100 and 200, random initial positions are generated. 
	The parameters adopted in this scenario are the same as those in Section \ref{sec_S2}. 
	The finial position and the trajectory of each robot are shown in Fig. \ref{S3} A. 
	One can see that our strategy exhibits stable performance in different shapes
	and swarm scales.
	
	Besides, to evaluate the performance of the proposed strategy, statistical simulation results are provided in Fig. \ref{S3} and the following three metrics are considered. 
	The coverage rate, entering rate and uniformity represent the proportion of the desired shape, the proportion of the robots that enter the desired shape and distribution uniformity of the robot swarms, respectively.
	The mathematical definitions of the three performance metrics can be found in our previous work \cite{Sun2023NC}, we omit them for brief.
	We conduct 20 statistical simulation experiments with the initial positions of the robot swarms being randomly generated. 
	Fig. \ref{S3} B shows the minimum, maximum, and average values of each metric in statistical simulation results. 
	Specifically, the coverage rate and entering rate remain 100$\%$ and the uniformity converges to a small value which means the robot swarms are evenly distributed in the desired shape. 
	One can see that the convergence time of the algorithm does not increase significantly with the swarm scale increasing, which verifies the efficiency of the proposed methods. 
	More additional simulation results can be found in Appendix \ref{additional_simulation} of this article.
	
	\section{Experiment Results}
	\label{Sec_experiment}
	In this section, indoor and outdoor experimental results are provided to evaluate the effectiveness of the proposed relative localization method and distributed shape formation strategy. 
	The video of the simulation and experiment can be found in \href{https://youtu.be/qMuPmzggDeU}{https://youtu.be/qMuPmzggDeU}.

	\subsection{Experiment Setup}
	\begin{figure*}[!t]\centering
		\includegraphics[scale=1.07]{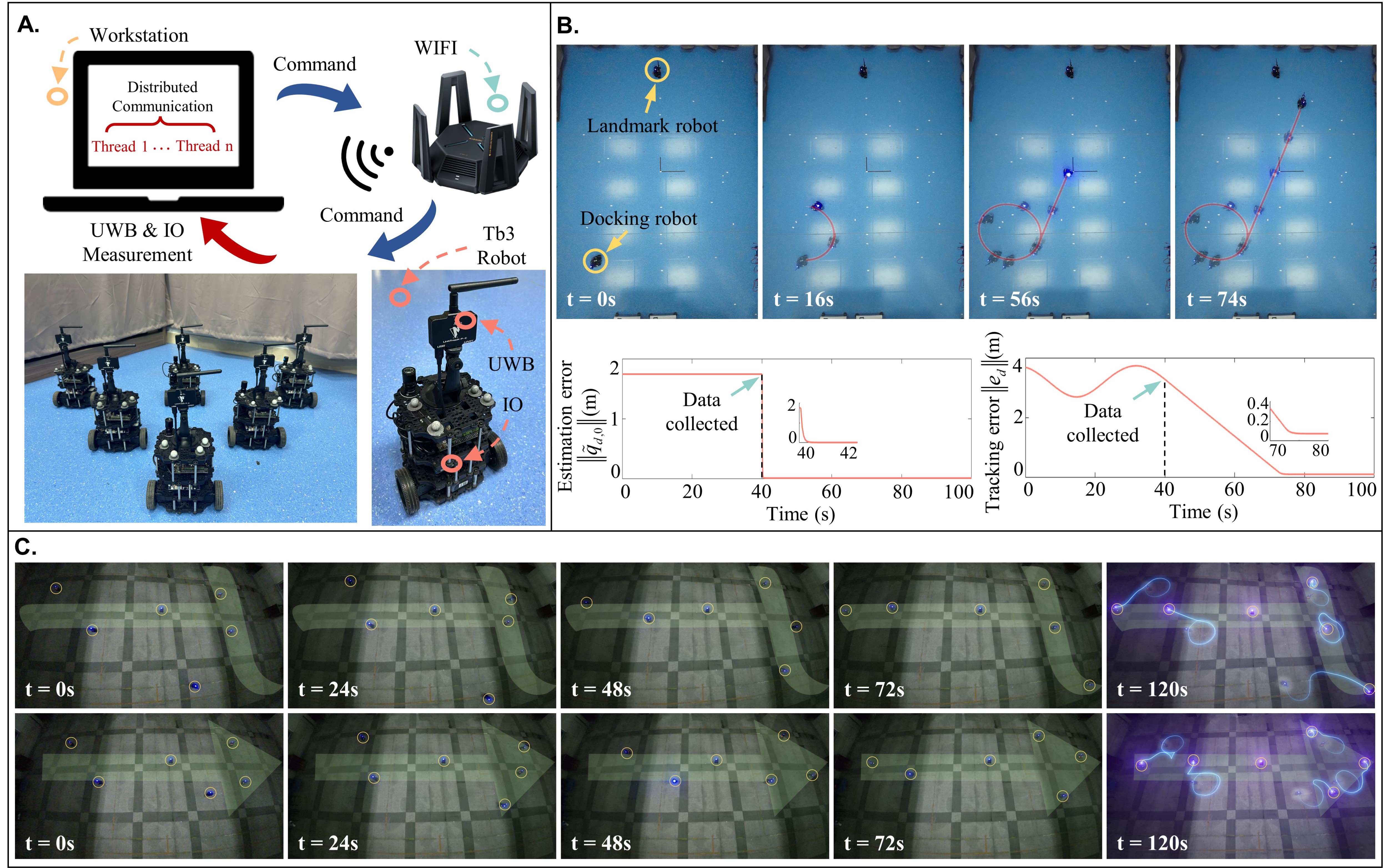}
		\caption{Indoor and outdoor experiment results \textbf{A}. Diagram of the experimental system. \textbf{B}. Indoor relative docking experiment results. Here, the estimation error $\tilde{q}_{d,0}$ and tracking error $e_{d}$ are defined as $\tilde{q}_{d,0}(t) = \hat{p}_{d,0}(t) - (p_{0} - p_{d}(t))$ and $e_{d}(t) = p_{d}(t) - p_{0} - p^{*}$. \textbf{C}. Outdoor shape formation experiment with a swarm consisted of 6 turtlebot3 robots.}\centering
		\label{exp}
		\vspace{-0.5cm}
	\end{figure*}

	The relative docking experiment is carried out in an indoor environment while the shape formation experiments are carried out in an outdoor environment. 
	In Fig. \ref{exp}A, the diagram of the experimental system is illustrated. 
	We use a kind of nonholonomic robot turtlebot3 as the experimental platform. 
	The method for transforming the omnidirectional control command into control command for nonholonomic robot can be found in \cite{Zhao2017TAC}. 
	At the sensor level, each robot is equipped with an IO to measure its own displacement and a UWB sensor to measure the real-time distance to the surrounding robots. 
	There are $n$ threads running on a ground station, each thread $i$ receives the real-time  measurement information sent by robot $i$. 
	Under the measurement topology defined in (\ref{neighbor}), distributed communication between threads are imitated with robot operating system (ROS). 
	Furthermore, the proposed relative localization and distributed shape formation control strategy for robot $i$ are also executed by thread $i$. 
	The robot swarms and ground station are in the same WiFi network to maintain the wireless communication between thread $i$ and robot $i$. 
	It should be noted that despite the usage of a ground station in the experiments, the algorithm still runs in a distributed manner. 
	
	\subsection{Relative Docking}
	
	In this experiment, a landmark robot is randomly placed in the experimental field, as shown in Fig. \ref{exp}B. 
	To verify the effectiveness of the proposed relative localization method, another robot named as docking robot is tasked to precisely move to the position $p*=[0.5,-0.5]^T$ relative to the landmark robot. 
	Denote the position of the landmark robot and docking robot as $p_{0}$ and $p_{d}(t)$.
	In the experiment, the control input of the docking robot is simply chosen as the following P-controller $v_{d}(t) = -\kappa(\hat{p}_{d,0}(t) - p^{*})$, where $\hat{p}_{d,0}(t)$ denotes the estimated relative position between the docking robot and landmark robot. 
	Other parameters are $\kappa=0.02$, $v_{\mathrm{max}}=0.1$m/s, $\Delta t=0.1$, $h=20$, $\lambda_{0}=0.2$. 
	
	In this experiment, NOKOV motion capture system is used to record the real position of the landmark robot and the docking robot.
	The trajectory of the docking robot is shown in Fig. \ref{exp}B. 
	The norm value of the estimation error $\tilde{q}_{d}(t)$ is shown in Fig. \ref{exp}B.
	One can see that when the measurement data is collected sufficiently, the estimation error fast converges into a small set near 0. 
	The norm value $\lVert e_{d}(t) \rVert$ of the tracking error is shown in Fig. \ref{exp}B. 
	The tracking error is around 10 cm, 
	which verifies the effectiveness of our localization algorithm.
	
	\subsection{Distributed Shape Formation}
	
	In the outdoor experiments, a swarm consisting of 6 turtlebot robots is assigned to form an ``arrow'' shape and a capital letter ``T'' in an outdoor environment range of $8\times13.5$ m$^2$. 
	The specific settings and parameter descriptions of the experiment are as follows.
	The nominal measurement range of the sensor is 500m, and the nominal ranging accuracy is 10cm. To simulate robots that can only obtain local information, we artificially limit the measurement range as $r_{\mathrm{sense}}=6$m. 
	To ensure stable communication between the two robots during relative positioning, $r_{\mathrm{neigh}}$ is chosen as $r_{\mathrm{neigh}}=4$m. Furthermore, collision avoidance is triggered if the inter-robot distance is less than $r_{\mathrm{avoid}}=2.5$m. 
	The parameters adopted in the relative localization and shape agreement are $\Delta t=0.05$s, $h=40$, $\lambda_{0}=0.2$, $c_{1}=0.1$, $\alpha=0.5$, $\delta_t = 1$s, $\delta_{0}=0.02$. 
	The parameters used in the behavior based control are $r_{i} = 0.5$m, $v_{max}=0.15$m/s, $w_{0}=0.32$rad/s, $w_{r}=0.05$rad/s, $\kappa_{1}=10$, $\kappa_{2}=15$, $\kappa_{3}=23$, $\kappa_{4}=2$. 
	
	In this experiment, the cell number of the two graphical binary graph shapes are $n_{\mathrm{cell, arrow}}=2209$ and $n_{\mathrm{cell, T}}=2266$. 
	One of the six robots is randomly selected as the seed robot and placed in the center of the experimental site. 
	As shown in the Fig. \ref{exp}A, each robot is equipped with a LED spotlight, and its motion trajectory is represented by the blue trajectory obtained from delay photography. 
	The whole process of the shape formation is shown in Fig. \ref{exp}C, the algorithm begins at $t=0$s, all the robots begin to move. At $t=24$s, robots move and collect measurement data. After the shape localization agreement, robots move to form the desired shape at $t=48$s. During the shape formation process, some robots may also localize the new neighboring robots. At $t=72$s, the shape formation process is almost complete. The final position of each robot at $t=120$s is represented by a purple highlight while the trajectories of all the robots are also shown as the blue curves in Fig. \ref{exp}C. 
	In fact, the actual experimental field is rugged and the robot does not have a shock absorption function, which may cause additional measurement errors for the IO. 
	Under these adverse conditions, the swarm consisting of 6 robots can still form the specified shape well.
	It proves the effectiveness, reliability, and robustness of the relative localization based shape formation algorithm.
	
	\section{Conclusion}
	\label{Sec_conclusion}
	
	This article has addressed the relative localization based shape formation problem for massive robot swarms.
	The motivation stems from applications such as object transport and building firefighting, where external localization infrastructures are unavailable.
	First, we proposed a concurrent-learning based relative position estimator, enabling inter-robot relative localization without the necessity of the well-known PE condition.
	Second, to achieve the shape localization agreement across the entire swarms in the absence of global coordinates, we proposed a finite-time consensus based agreement protocol.
	Third, we devised a novel behavior-based shape formation control strategy that can better exploit the historical measurement information to enhance the observability of relative localization among robots. 
	This control strategy has a concise structure and can be implemented on robotic systems easily.
	Extensive simulation results and comparisons with state-of-the-art relative localization methods were presented.
	Additionally, outdoor experiments involving up to six robots were conducted to verify the robustness, effectiveness and adaptability of the proposed methods.
	

	
	\appendices
		\section{Theoretical Analysis}
			\subsection{Preliminary lemmas}
		\label{P_l}
		The following lemmas are useful to establish main results.
		
		\begin{lemma} \cite{Biggs1993book}
			\label{graph}
			If the undirected graph $\mathcal{G}$ is connected and at least one agent has direct access to the seed robot. The matrix $(\mathcal{L} + \mathcal{B})$ is nonsingular.
		\end{lemma}
		
		\begin{lemma} \cite{Beckenbach2012book}
			\label{inequality1}
			For $p\in \mathbb{R}$, $q\in \mathbb{R}$ and $0 < \alpha < 1$, if $p \geq q$, the following inequality holds:
			\begin{align}
				\lvert p - q \rvert^{\alpha} \geq \lvert p 	\rvert^{\alpha} - 2\lvert q \rvert^{\alpha}.
				\notag
			\end{align}
			Furthermore, if $p \geq q \geq 0$, the following inequality holds:
			\begin{align}
				\lvert p - q \rvert^{\alpha} \leq 2\lvert p \rvert^{\alpha} - \lvert q \rvert^{\alpha}.
				\notag
			\end{align}
		\end{lemma}
		
		\begin{lemma} \cite{Beckenbach2012book}
			\label{inequality2}
			For any vector $x = [x_{1},...,x_{n}]^T\in \mathbb{R}^n$, the $p$-norm of $x$ is defined as $\lVert x \rVert_{p} = \left( \sum_{i=1}^{n} |x_{i}|^p \right)^\frac{1}{p}$. For the constants $p$, $q$, if $p \geq q \geq 1$, the following inequality holds:
			\begin{align}
				\notag
				\lVert x \rVert_{p} \leq \lVert x \rVert_{q} \leq n^{\frac{1}{q}-\frac{1}{p}} \lVert x \rVert_{p}.
			\end{align}
		\end{lemma}
		
		\begin{lemma}
			\label{inequality3}
			For $p\in \mathbb{R}^{n}$, $q\in \mathbb{R}^{n}$ and $0 < \alpha < 1$, the following inequality holds:
			\begin{align}
				\notag
				-p^{T} \mathrm{sig}(p-q)^{\alpha} \leq - \frac{\alpha}{1+\alpha} \lVert p \rVert ^{1+\alpha} + \frac{\alpha}{1+\alpha} 2^{\frac{{1+\alpha}}{\alpha}} n^{\frac{1-\alpha}{2(\alpha+1)}} \lVert q \rVert_{1}^{1+\alpha}.
			\end{align}
		\end{lemma}
		\begin{IEEEproof}
			Let $p = [p_{1},...,p_{n}]^T$ and $q = [q_{1},...,q_{n}]^T$. Obviously, it has 
			\begin{align}
				\label{lemma_4_1}
				-p^{T} \mathrm{sig}(p-q)^{\alpha} = \sum_{i=1}^{n} -p_{i} \mathrm{sig}(p_{i} - q_{i})^{\alpha}.
			\end{align}
			
			Define $\delta_{i} \triangleq -p_{i} \mathrm{sig}(p_{i}-q_{i})^{\alpha}$ for brief, the following different situations will be discussed,
			
			\textit{Case 1}: $p_{i} \geq 0$, $p_{i} \geq q_{i}$, according to Lemma \ref{inequality1}, it has
			\begin{align}
				\notag
				\delta_{i} = - \lvert p_{i} \rvert \lvert p_{i} - q_{i} \rvert^{\alpha} 
				\leq - \lvert p_{i} \rvert \left( \lvert p_{i} \rvert^{\alpha} - 2\lvert q_{i} \rvert^{\alpha} \right).
			\end{align}
			
			\textit{Case 2}: $p_{i} \geq 0$, $p_{i} < q_{i}$, according to Lemma \ref{inequality1}, it has
			\begin{align}
				\notag
				\delta_{i} &= \lvert p_{i} \rvert \lvert q_{i} - p_{i} \rvert^{\alpha} \leq \lvert p_{i} \rvert \left( 2\lvert q_{i} \rvert^{\alpha} - \lvert p_{i} \rvert^{\alpha} \right)\\
				\notag
				&= - \lvert p_{i} \rvert \left( \lvert p_{i} \rvert^{\alpha} - 2\lvert q_{i} \rvert^{\alpha} \right).
			\end{align}
			
			\textit{Case 3}: $p_{i} < 0$, $p_{i} \geq q_{i}$, according to Lemma \ref{inequality1}, it has
			\begin{align}
				\notag
				\delta_{i} &= \lvert p_{i} \rvert \lvert |q_{i}| - |p_{i}| \rvert^{\alpha} \leq \lvert p_{i} \rvert \left( 2\lvert q_{i} \rvert^{\alpha} - \lvert p_{i} \rvert^{\alpha} \right)\\
				\notag
				&= - \lvert p_{i} \rvert \left( \lvert p_{i} \rvert^{\alpha} - 2\lvert q_{i} \rvert^{\alpha} \right).
			\end{align}
			
			\textit{Case 4}: $p_{i} < 0$, $p_{i} < q_{i}$, consider the two sub-cases:
			
			(i). When $q_{i}<0$ according to Lemma \ref{inequality1}, it has
			\begin{align}
				\notag
				\delta_{i} &= -\lvert p_{i} \rvert \lvert |p_{i}| - |q_{i}| \rvert^{\alpha} \leq -\lvert p_{i} \rvert \left( \lvert p_{i} \rvert^{\alpha} - 2\lvert q_{i} \rvert^{\alpha} \right).
			\end{align}
			
			(ii). When $q_{i} \geq 0$, one can see that
			\begin{align}
				\notag
				\delta_{i} &= -\lvert p_{i} \rvert \lvert |p_{i}| + |q_{i}| \rvert^{\alpha} \leq -\lvert p_{i} \rvert \left( \lvert p_{i} \rvert^{\alpha} - 2\lvert q_{i} \rvert^{\alpha} \right).
			\end{align}
			
			As a result, the following inequality holds
			\begin{align}
				\label{lemma_4_2}
				-p_{i} \mathrm{sig}(p_{i}-q_{i}) \leq - \lvert p_{i} \rvert \left( \lvert p_{i} \rvert^{\alpha} - 2\lvert q_{i} \rvert^{\alpha} \right).
			\end{align}
			According to Young's inequality $mn \leq \frac{m^l}{l} + \frac{n^k}{k}$, $l,k > 0$ and $\frac{1}{l}+\frac{1}{k} = 1$. Take $l=\alpha+1$, it has
			\begin{align}
				\label{lemma_4_3}
				\lvert p_{i} \rvert \left( 2\lvert q_{i} \rvert^{\alpha} \right) \leq \frac{1}{1+\alpha} \lvert p_{i} \rvert^{1+\alpha} +\frac{\alpha}{1+\alpha} 2^{\frac{{1+\alpha}}{\alpha}} \lvert q_{i} \rvert^{1+\alpha}.
			\end{align}
			Substituting (\ref{lemma_4_2}), (\ref{lemma_4_3}) int (\ref{lemma_4_1}), it has
			\begin{align}
				\notag
				-p^T \mathrm{sig}(p - q)^{\alpha} \leq - \frac{\alpha}{1+\alpha} \lVert p \rVert ^{{1+\alpha}} + \frac{\alpha}{1+\alpha} 2^{\frac{{1+\alpha}}{\alpha}} n^{\frac{1-\alpha}{2(\alpha+1)}} \lVert q \rVert^{1+\alpha}.
			\end{align}
		\end{IEEEproof}
		
		\begin{lemma} \cite{Zhu2011IJRNC}
			\label{finite_time}
			Consider the system $\dot{x} = f(x,u)$. Suppose that there exist continuous function $V(t) : [0,\infty) \rightarrow [0,\infty)$, scalars $K>0$, $0<\alpha<1$ and $0<\beta<\infty$ such that
			\begin{align}
				\notag
				\dot{V}(x) \leq -K V(x)^{\alpha} + \beta.
			\end{align}
			Then the state $x$ of the closed-loop system converges into the set $x \in \left\{ V(x)^{\alpha} \leq \frac{\beta}{(1-\gamma)K} \right\}$ in finite time $t^*$, where $t^*$ is
			\begin{align}
				\notag
				t^* = \frac{V(x(t_{0}))^{(1-\alpha)}}{K\gamma(1-\alpha)}.
			\end{align}
		\end{lemma}
		
		\begin{lemma}
			\label{eigenvalue}
			For symmetric matrices $M= \begin{bmatrix}
				a_{1}&a_{3} \\
				a_{3}&a_{2}
			\end{bmatrix}$, where $a_{1}>0$, $a_{2}>0$ and $a_{3}\in \mathbb{R}$. Then the maximum value of the $\frac{\lambda_{\min}(M)}{\lambda_{\min}(M)}=\frac{\min\{ a_{1},a_{2} \}}{\max\{ a_{1},a_{2} \}}$, if and only if $a_{3}=0$.
		\end{lemma}
		\begin{IEEEproof}
			One can calculate that $\lambda_{\min}(M), \lambda_{\max}(M) = \frac{a_{1}+a_{2} \pm \sqrt{(a_{1}-a_{2})^2 + 4(a_{3})^2}}{3}.$
			One can see that $\frac{\lambda_{\min}(M)}{\lambda_{\min}(M)}$ is a monotonically decreasing function about $(a_{3})^2$, $\frac{\lambda_{\min}(M)}{\lambda_{\min}(M)}$ take the maximum value $\frac{\min\{ a_{1},a_{2} \}}{\max\{ a_{1},a_{2} \}}$ at $a_{3} = 0$.
		\end{IEEEproof}
		
		\subsection{Proof of Theorem 1}
		\label{P_T1}
		The dynamics of the relative localization error is
		\begin{align}
			\label{T_1}
			\tilde{p}_{ij,0}(t_{k+1}) = \left(I - \eta U_{ij}(t_{k}) - \eta S_{ij} \right) \tilde{p}_{ij,0}(t_{k}).
		\end{align}
		Choose a Lyapunov candidate as $V_{a}(t_{k}) = \tilde{p}_{ij,0}^T(t_{k})\tilde{p}_{ij,0}(t_{k}).$
		Define $\Delta V_{a} \triangleq V_{a}(t_{k+1}) -V_{a}(t_{k})$, then it has,
		\begin{align}
			\label{T_3}
			\Delta V_{a} 
			\notag
			= &\tilde{p}_{ij,0}^T(t_{k}) \left(-2\eta S_{ij} + 2\eta^2S_{ij}U_{ij}(t_{k}) + \eta^2S_{ij}^2 \right) \tilde{p}_{ij,0}(t_{k}) \\
			& - \tilde{p}_{ij,0}^T(t_{k}) \eta U_{ij}(t_{k}) (2I - \eta U_{ij}(t_{k}) ) \tilde{p}_{ij,0}(t_{k}).
		\end{align}
		It can be checked that
		\begin{subequations}
			\label{T_4}
			\begin{align}
				\lambda_{\min}(U_{ij}(t_{k})) \lVert \tilde{p}_{ij,0}(t_{k}) \rVert \leq \lVert U_{ij}  \tilde{p}_{ij,0}(t_{k}) \rVert \\
				\lambda_{\max}(U_{ij}(t_{k})) \lVert \tilde{p}_{ij,0}(t_{k}) \rVert \geq \lVert U_{ij}  \tilde{p}_{ij,0}(t_{k}) \rVert.
			\end{align}
		\end{subequations} 
		Similar to $S_{ij}$, substituting (\ref{T_4}) in (\ref{T_3}), it has
		\begin{align}
			\notag
			\Delta V_{a} &\leq \lVert \tilde{p}_{ij,0}(t_{k}) \rVert^2\left(-2 \eta \lambda_{\min}(S_{ij}) + 2\eta^2 \lambda_{\max}(U_{ij}(t_{k})) \lambda_{\max}(S_{ij}) \right. \\
			\notag
			& \left. + \eta^2 \lambda_{\max}(S_{ij})^2 - 2 \eta \lambda_{\min}(U_{ij}(t_{k})) + \eta^2 \lambda_{\max}(U_{ij}(t_{k}))^2 \right).
		\end{align}
		With the learning rate $\eta$ defined in (\ref{Equ_step}), it has
		\begin{align}
			\notag
			\Delta V_{a} \leq & \frac{-\lambda_{\min}(S_{ij})^2}{(\lambda_{\max}(U_{ij}(t_{k}))(t_{k}) + \lambda_{\max}(S_{ij})) ^2} \lVert \tilde{p}_{ij,0}(t_{k}) \rVert^2.
		\end{align}
		
		One can see that the maximum eigenvalue of $U_{ij}(t_{k})$ satisfies $\lambda_{\max}(U_{ij}(t_{k})) = \lVert u_{ij}(t_{k}) \rVert \leq 2v_{\max}\Delta t$. Then it has,
		\begin{align}
			\notag
			V_{a}(t_{k}) \leq \left( 1-\frac{\lambda_{\min}(S_{ij})^2}{(2v_{\max}\Delta t + \lambda_{\max}(S_{ij})) ^2} \right)^k V_{a}(t_{0}).
		\end{align}
		As a result, the relative localization error $\tilde{p}_{ij,0}$ satisfies 
		\begin{align}
			\notag
			\lVert \tilde{p}_{ij,0}(t_{k}) \rVert \leq \left( \lambda_{ij} \right)^{k} \lVert \tilde{p}_{ij,0}(t_{0}) \rVert
		\end{align} 
		where $\lambda_{ij}$ is defined in (\ref{Equ_converge_rate})
		
		\subsection{Proof of Theorem 2}
		\label{P_T2}
		Define the recorded relative displacement data $u_{ij}(t_{{\rm c},m}) \triangleq [u_{x,ij}(t_{{\rm c},m}), u_{y,ij}(t_{{\rm c},m})]^T$, as a result, the data matrix $S_{ij}$ is $S_{ij} =\begin{bmatrix}
			s_1 & s_3 \\
			s_3 & s_2
		\end{bmatrix}$,
		where $s_1$, $s_2$ and $s_3$ are defined as $s_1=\sum_{m=1}^{\varsigma_{0}ij} u_{x,ij}(t_{{\rm c},m})^2$, $s_2=\sum_{m=1}^{\varsigma_{0}ij} u_{y,ij}(t_{{\rm c},m})^2$ and $s_3=\sum_{m=1}^{\varsigma_{0}ij} u_{x,ij}(t_{{\rm c},m})u_{y,ij}(t_{{\rm c},m})$. According to Theorem \ref{Theorem 1} and Lemma \ref{eigenvalue}, the maximum converge rate of the proposed estimator can be achieved if and only if $s_3 = 0$ and $s_1 = s_2 \neq 0$.   
		When robot $i$ and robot $j$ perform the designed motion (\ref{v_ij}), the relative velocity between them is $v_{ij}(t) = [(r_{i}\mathrm{cos}(w_{i}t) - r_{j}\mathrm{cos}(w_{j}t)), (r_{i}\mathrm{sin}(w_{i}t) - r_{j}\mathrm{sin}(w_{j}t))]^T$.
		Before giving the main proof, we first present several useful equations.
		With condition (\ref{con_num_interval}) is satisfied, one can see that $t_{{\rm c},m} = \frac{2\pi}{\varsigma_{0}}(m-1)$.
		Recall the angular velocity $w_{i} =1/i$ of robot $i$, then it has
		\begin{align}
			\label{T2_1_5}
				&2\sum_{m=1}^{\varsigma_{0}ij} \mathrm{sin}(w_i t_{{\rm c},m})\mathrm{sin}(w_jt_{{\rm c},m}) \\
				\notag
				= & \sum_{m=1}^{\varsigma_{0}ij} \left( \mathrm{cos}(w_i t_{{\rm c},m} - w_j t_{{\rm c},m}) - \mathrm{cos}(w_i t_{{\rm c},m} + w_j t_{{\rm c},m}) \right) \\
				\notag
				= & \sum_{m=0}^{\varsigma_{0}ij-1} \left( \mathrm{cos}\left( \frac{2\pi(i-j)m}{\varsigma_{0}ij} \right) - \mathrm{cos}\left( \frac{2\pi(i+j)m}{\varsigma_{0}ij} \right) \right). 
		\end{align}
		
		It can be seen that (\ref{T2_1_5}) is equal to zero. 
		One can check that the following equation holds when condition (\ref{con_num_interval}) is satisfied,
		\begin{subequations}
			\label{T2_2}
			\begin{align}
				&\sum_{m=1}^{\varsigma_{0}ij} f(w_i t_{{\rm c},m})f(w_jt_{{\rm c},m})=0, f \in \{ \mathrm{sin}, \mathrm{cos} \} \\
				&\sum_{m=1}^{\varsigma_{0}ij} \mathrm{sin}(w_1 t_{{\rm c},m})\mathrm{cos}(w_2 t_{{\rm c},m}) = 0, w_1,w_2 \in \{ w_{i}, w_{j} \} \\
				&\sum_{m=1}^{\varsigma_{0}ij} \mathrm{sin}(w_i t_{{\rm c},m})^2 = \sum_{m=1}^{M} \mathrm{cos}(w_i t_{{\rm c},m})^2 \neq 0.
			\end{align}
		\end{subequations}
		
		Besides, consider the fact that
		\begin{subequations}
			\label{T2_3}
			\begin{align}
				\notag
				u_{x,i}(t_{{\rm c},m}) &= 2r_{i}\mathrm{sin}\left( \frac{w_{i}h\Delta t}{2} \right) \mathrm{cos}\left( w_{i}\left( \frac{h\Delta t}{2}+t_{{\rm c},m} \right) \right) \\
				&= s_{i}\mathrm{cos}(w_{i}t_{{\rm c},m}) - c_{i}\mathrm{sin}(w_{i}t_{{\rm c},m}) \\
				u_{y,i}(t_{{\rm c},m}) &= s_{i}\mathrm{sin}(w_{i}t_{{\rm c},m}) + c_{i}\mathrm{cos}(w_{i}t_{{\rm c},m})
			\end{align}
		\end{subequations}
		where $s_{i}=r_{i}\mathrm{sin}(w_{i}h\Delta t)$ and $c_{i}=1-r_{i}\mathrm{cos}(w_{i}h\Delta t)$. According to (\ref{T2_2}) and (\ref{T2_3}), it can be checked that
		\begin{align}
			\notag
			s_3 = &\sum_{m=1}^{\varsigma_{0}ij} \left( s_{i}\mathrm{cos}(w_{i}t_{{\rm c},m}) - c_{i}\mathrm{sin}(w_{i}t_{{\rm c},m}) - s_{j}\mathrm{cos}(w_{j}t_{{\rm c},m}) \right.\\
			\notag
			&\left. + c_{j}\mathrm{sin}(w_{j}t_{{\rm c},m}) \right) \left( s_{i}\mathrm{sin}(w_{i}t_{{\rm c},m}) + c_{i}\mathrm{cos}(w_{i}t_{{\rm c},m})\right.\\
			\notag
			&\left. - s_{j}\mathrm{sin}(w_{j}t_{{\rm c},m}) - c_{j}\mathrm{cos}(w_{j}t_{{\rm c},m}) \right) = 0 \\
			\notag
			s_{1} =& \sum_{m=1}^{\varsigma_{0}ij} \left( s_{i}\mathrm{cos}(w_{i}t_{{\rm c},m}) - c_{i}\mathrm{sin}(w_{i}t_{{\rm c},m}) - s_{j}\mathrm{cos}(w_{j}t_{{\rm c},m}) \right.\\
			\notag
			&\left. + c_{j}\mathrm{sin}(w_{j}t_{{\rm c},m}) \right)^2 = \sum_{m=1}^{\varsigma_{0}ij} \left( s_{i}\mathrm{sin}(w_{i}t_{{\rm c},m}) + c_{i}\mathrm{cos}(w_{i}t_{{\rm c},m}) \right.\\
			\notag
			&\left. - s_{j}\mathrm{sin}(w_{j}t_{{\rm c},m}) - c_{j}\mathrm{cos}(w_{j}t_{{\rm c},m}) \right)^2 = s_{2} \neq 0
		\end{align} 
		which means the proposed observer (\ref{Equ_observer}) converges the fastest with the recorded data measurements satisfying condition (\ref{con_num_interval}).
		
		\subsection{Proof of Theorem 3}
		\label{P_T3}
		For each robot $j$ in robot $i$'s initial neighbor set $\mathcal{N}_{i}(t_{0})$, according to the definition of the neighboring robot, the initial distance $d_{ij}(t_{0})$ must satisfy $d_{ij}(t_{0})<r_{\mathrm{neigh}}$. 
		When robot $i$ makes circular motion with radius $r_{i}$, the maximum distance $d_{ij, \max}$ between them satisfies $d_{ij,\max} \leq d_{ij}(t_{0}) + r_{i} + r_{j} < r_{\mathrm{sense}}$. 
		Robot $j$ is still a neighbor to robot $i$. As a result, the initial neighbor set $\mathcal{N}_{i}(t_{0})$ remains robot $i$'s neighbor.
		
		For each robot $i$ which is in the sense range of the seed robot initially, i.e. $\mu_{i} = 1$, recall the direct estimation error $\tilde{p}_{i0,0} = \hat{p}_{i0,0} - p_{i0}(t_{0})$ and the agreement error $\tilde{q}_{i,0} = \hat{q}_{i,0} - p_{i0}(t_{0})$.
		As a result, it has
		\begin{align}
			\label{T3_1}
			\notag
			\dot{\tilde{q}}_{i,0}(t) & = -c_{1} \mathrm{sig} \Big( \sum_{j=1}^n a_{ij} \left( \tilde{q}_{i,0}(t) - \tilde{q}_{j,0}(t) -\tilde{p}_{ij,0}(t) \right) \\
			& + \mu_{i} \left( \tilde{q}_{i,0}(t) - \tilde{p}_{i0,0}(t) \right) \Big)^{\alpha}.
		\end{align}
		Define $\chi_{i} = \sum_{j=1}^n a_{ij} \left( \tilde{q}_{i,0}(t) - \tilde{q}_{j,0}(t) + \mu_{i} \tilde{q}_{i,0}(t) \right) \in \mathbb{R}^2 $ and $\chi = [\chi_{1},...,\chi_{N}]^T$, then it has
		\begin{align}
			\label{T3_2}
			\sum_{i=1}^{n}\chi_{i}^2 = \tilde{q}^T(\mathcal{L} + \mathcal{B})^T(\mathcal{L} + \mathcal{B})\tilde{q} \leq \lambda_{\min}(\mathcal{L} + \mathcal{B}) \tilde{q}^T (\mathcal{L} + \mathcal{B}) \tilde{q}.
		\end{align}
		Substituting (\ref{T3_2}) in (\ref{V_l}), one can see that
		\begin{align}
			\label{T3_3}
			V_{l} \leq \frac{1}{\lambda_{\min}(\mathcal{L} + \mathcal{B})}\sum_{i=1}^{n} \chi_{i}^2.
		\end{align}
		
		According to lemma \ref{graph}, $\lambda_{\min}(\mathcal{L} + \mathcal{B}) > 0$. According to (\ref{T3_1}), the derivative of $V_{l}$ is derived as
		\begin{align}
			\label{T3_4}
			\dot{V}_{l} = \tilde{q}^T (\mathcal{L} + \mathcal{B}) \dot{\tilde{q}} 
			= - c_{1} \sum_{i=1}^{n} \chi_{i}^T \mathrm{sig} \left( \chi_{i} - \delta_{i} \right)^{\alpha}
		\end{align}
		where $\delta{i} \triangleq \sum_{j=1}^n a_{ij}\tilde{p}_{ij,0}(t) + \mu_{i} \tilde{p}_{i0,0}(t) \in \mathbb{R}^2$. According to Lemma \ref{inequality3}, (\ref{T3_4}) can be simplified as
		\begin{align}
			\notag
			\dot{V}_{l} \leq \sum_{i=1}^{n} \left( - \frac{\alpha}{1+\alpha} \lVert \chi_{i} \rVert^{1 + \alpha} + 
			\frac{\alpha}{1+\alpha} 2^{\frac{1+3\alpha}{2\alpha(\alpha+1)}} \lVert \delta_{i} \rVert^{1+\alpha} \right).
		\end{align}
		According to Eq. (\ref{T3_3}), it has
		\begin{align}
				\notag
				\dot{V}_{l} & \leq -\frac{\alpha}{1+\alpha}\left( \sum_{i=1}^{n} \lVert \chi_{i} \rVert^2 \right)^{\frac{1+\alpha}{2}} + \frac{\alpha}{1+\alpha} 2^{\frac{1+3\alpha}{2\alpha(\alpha+1)}}  \sum_{i=1}^{n} \lVert \delta_{i} \rVert^{1+\alpha} \\
				\notag
				& \leq -\frac{\alpha}{1+\alpha} \lambda_{\min}(\mathcal{L} + \mathcal{B}) ^{\frac{1+\alpha}{2}} V_{l}^{\frac{1+\alpha}{2}} 
				\\
				\notag
				& + \frac{\alpha}{1+\alpha} 2^{\frac{1+3\alpha}{2\alpha(\alpha+1)}} \left( \sum_{i=1}^{n} \lVert \delta_{i} \rVert^{1+\alpha} \right).
		\end{align}
		
		According to the definition of $\delta_{i}$ and Theorem \ref{Theorem 1}, one can see that
		$\lVert \delta_{i} \rVert$ is bounded for all $t>0$. As a result, $V_{l}$ is bounded.
		Furthermore, for agent $i$, $\tilde{p}_{ij,0}$ will converge to satisfy $\lVert \tilde{p}_{ij,0} \rVert \leq \varepsilon $ near zero in a finite time $t_{a,ij}$, where $\varepsilon>0$ is a constant and $t_{a, ij}$ can be given as
		\begin{align}
			\label{T3_6}
			t_{a, ij} = \frac{\mathrm{ln} (\varepsilon \lVert \tilde{p}_{ij,0}(t_{0}) \rVert^{-1})}{\mathrm{ln}(\lambda_{ij})} \Delta t.
		\end{align}
		
		Besides, with the condition that the measurement data matrix $S_{ij}$ is full rank for each pair $(i,j) \in \mathcal{G}(t_{0})$, $\lambda_{ij}$ is positive and $t_{a,ij}$ defined in (\ref{T3_6}) is finite. As a result, there exist a $t_{a,i}^{\mathrm{\max}} = \max\{ t_{a,ij},j \in \mathcal{N}_{i} \}$ such that when $t>t_{a,i}^{\mathrm{\max}}$, the following inequality holds:
		\begin{align}
			\notag
			\delta_{i} \leq |\mathcal{N}_{i}| \varepsilon
		\end{align} 
		where $|\mathcal{N}_{i}|$ denotes the initial neighbor number of robot $i$. As a result, there also exist a constant $t_{a} = \max\{ t_{a,i}^{\mathrm{\max}},i \in \mathcal{V} \}$ defined as (\ref{t_a}) such that when $t>t_{a}$, the following inequality holds:
		\begin{align}
			\label{T3_7}
			\sum_{i=1}^{n} \lVert \delta_{i} \rVert^{1+\alpha} \leq  \sum_{i=1}^{n} (|\mathcal{N}_{i}|\varepsilon)^{1+\alpha}.
		\end{align}
		
		Note that the right side of (\ref{T3_7}) is a constant. According to Lemma \ref{finite_time}, $V_{l}$ will converge into
		\begin{align}
			\label{T3_8}
			V_{l} \leq \frac{\frac{\alpha}{1+\alpha} 2^{\frac{1+3\alpha}{2\alpha(\alpha+1)}} \left( \sum_{i=1}^{n} (|\mathcal{N}_{i}|\varepsilon)^{1+\alpha} \right)}{(1-\gamma)\frac{\alpha}{1+\alpha} \lambda_{\min}(\mathcal{L} + \mathcal{B})}
		\end{align}
		for $t>t_{a} + t_{l}$, where $t_{l}$ is defined as (\ref{t_l}).
		According to (\ref{T3_8}) and the definition (\ref{V_l}) of $V_{l}$, one can get that the global estimation error $\tilde{q}$ will converge into
		\begin{align}
			\notag
			\lVert \tilde{q} \rVert \leq b
		\end{align}
		where $b$ is defined in ($\ref{b}$).
	
	\section{Technical Details}
	\subsection{Additional simulation results}
	\label{additional_simulation}
	\begin{figure}\centering
		\includegraphics[scale=1.045]{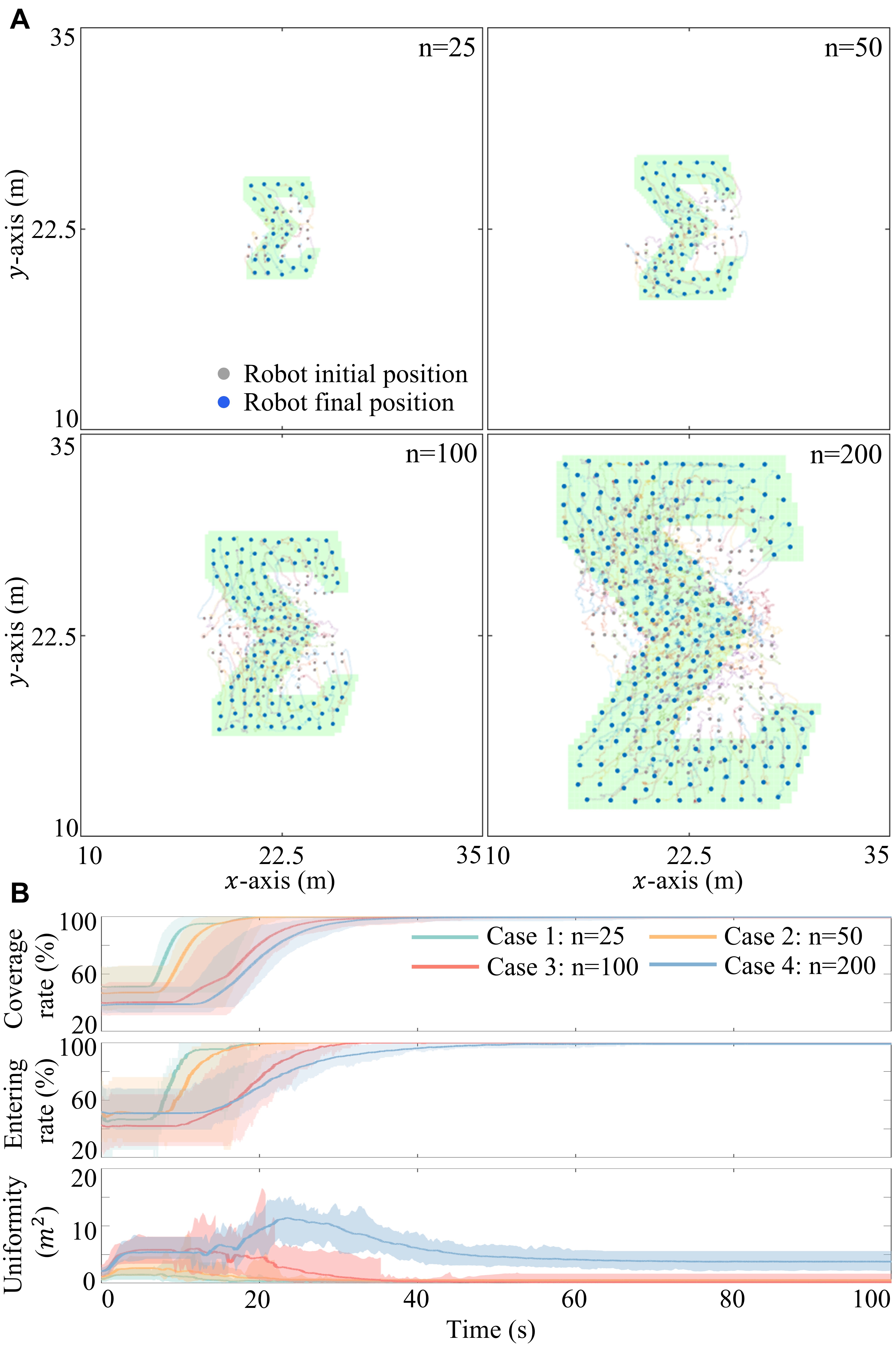}
		\caption{A. Trajectory and final position of the swarm with 25, 50, 100 and 200 robots.
		B. Coverage rate, entering rate and distribution uniformity of the robot swarms to form the mathematical sum symbol.
		}\centering
		\label{S3_1}
		\vspace{-0.5cm}
	\end{figure}

	\begin{figure}\centering
		\includegraphics[scale=1.045]{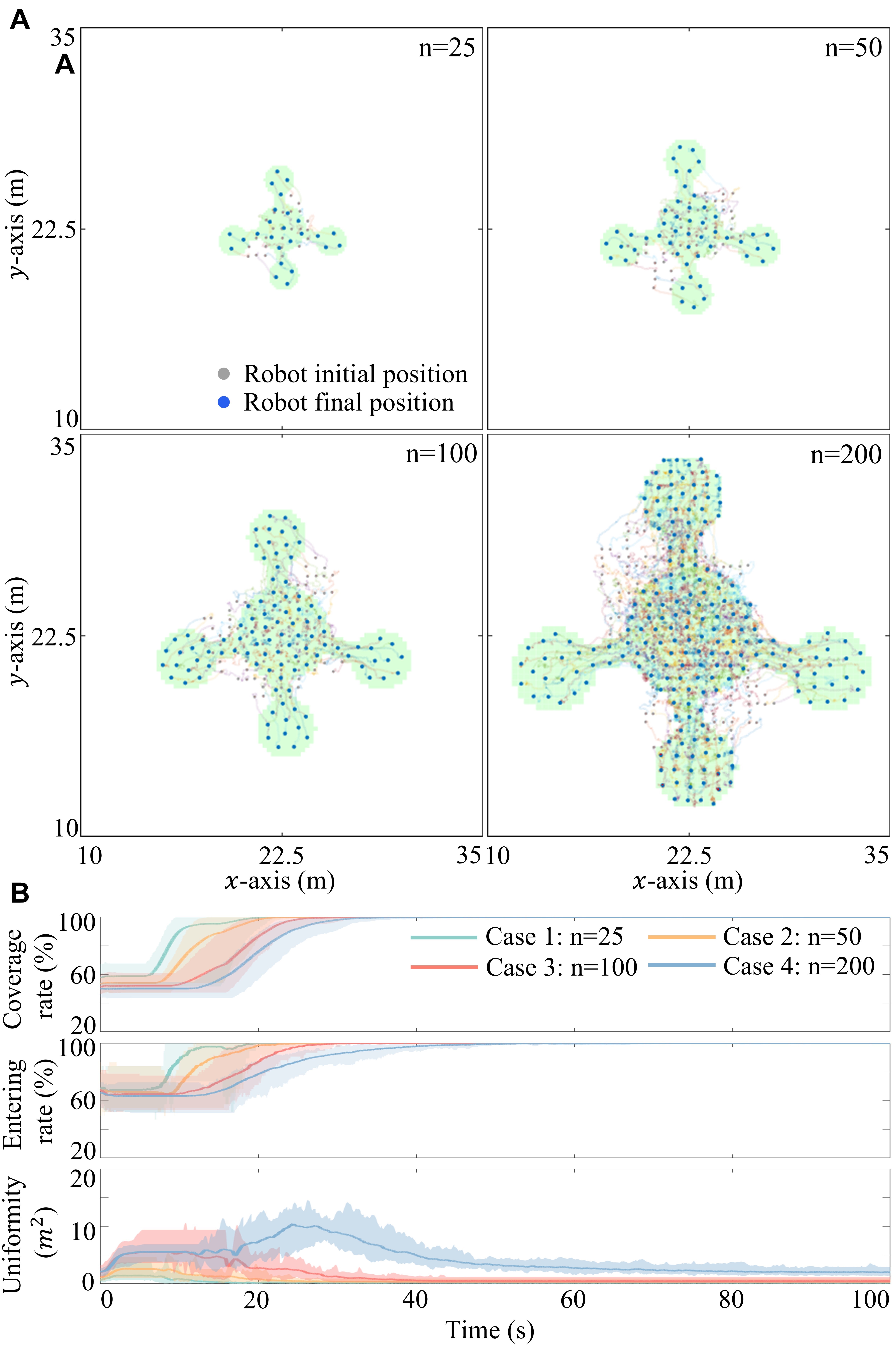}
		\caption{A. Trajectory and final position of the swarm with 25, 50, 100 and 200 robots.
		B. Coverage rate, entering rate and distribution uniformity of the robot swarms to form the methane molecule shape.
		}\centering
		\label{S3_2}
		\vspace{-0.5cm}
	\end{figure}
	In the adaptability test scenario, robot swarms with with scales 25, 50, 100 and 200 are assigned to form different shapes including the mathematical sum symbol and the methane molecule shape.
	Parameters adopted in this scenario are the same as those in Section \ref{sec_S2}. 
	The final position and the trajectory of each robot are shown in Fig. \ref{S3_1} A and Fig. \ref{S3_2} A. 
	One can see that our strategy exhibits stable performance in different shapes and swarm scales. Besides, we conduct 20 statistical simulation experiments with the initial positions of the robot swarms being randomly generated. 
	The scales of the robot swarm are 25, 50, 100 and 200 and the desired formation shapes are mathematical sum symbol and the methane molecule shape.
	Fig. \ref{S3_1} B and Fig. \ref{S3_2} B show the minimum, maximum, and average values of each metric in statistical simulation results. 
	Similarly, the coverage rate and entering rate remain 100$\%$ and the uniformity converges to a small value which means the robot swarms are evenly distributed in the desired shape, which verifies the efficiency of the proposed methods.
		
	\subsection{Discussion on parameter setting}
	\label{parameter_discussion}
	\begin{table*}[!t]
		\centering
		\caption{Parameter list}
		\begin{tabular}{m{4cm}m{0.7cm}m{0.7cm}m{0.7cm}m{0.7cm}m{0.7cm}m{0.7cm}m{0.7cm}m{0.7cm}m{0.7cm}cm{0.7cm}m{0.7cm}} 
			\toprule 
			Simulation \& Experiment & $\Delta t$(s) & $h$ & $\lambda_{0}$ & $c_{1}$  & $\alpha$ & $\delta_{t}$(s) & $\delta_{0}$(m) & $n_{\mathrm{cell}}$ & $\kappa_{1}$ & $\kappa_{2}$ & $\kappa_{3}$ & $\kappa_{4}$  \\ 
			\midrule 
			\textbf{Simulation}: dark shape &0.05 & 40 & 0.1 & 0.1 & 0.5 & 1 & 0.01 & 2529 & 10 & 15 & 25 & 2 \\
			\midrule 
			\textbf{Simulation}: letter ``R'' shape &0.05 & 40 & 0.1 & 0.1 & 0.5 & 1 & 0.01 & 2470 & 10 & 15 & 25 & 2 \\
			\midrule 
				\textbf{Simulation}: ``sum'' symbol shape &0.05 & 40 & 0.1 & 0.1 & 0.5 & 1 & 0.01 & 2130 & 10 & 15 & 25 & 2 \\
			\midrule 
				\textbf{Simulation}: ``molecule'' shape &0.05 & 40 & 0.1 & 0.1 & 0.5 & 1 & 0.01 & 2251 & 10 & 15 & 25 & 2 \\
			\midrule 
				\textbf{Experiment}: ``arrow'' shape &0.05 & 40 & 0.2 & 0.1 & 0.5 & 1 & 0.02 & 2209 & 10 & 15 & 23 & 2 \\
			\midrule 
				\textbf{Experiment}: letter ``T'' shape &0.05 & 40 & 0.2 & 0.1 & 0.5 & 1 & 0.02 & 2266 & 10 & 15 & 23 & 2 \\
			\bottomrule 
		\end{tabular}\label{parameter}
		\end{table*}
	We explain in detail all the parameters involved in the method proposed in this article and their corresponding meanings below. In addition, for all our simulation and hardware experimental scenarios, we have discussed the parameter selection method and the sensitivity of the proposed method to the parameters. 
	The main parameters adopted in our methods can be classified into two classes, one class is the relative localization and shape agreement related, including $\Delta t$, $h$, $\lambda_{0}$, $c_{1}$, $\alpha$, $\delta_{t}$ and $\delta_{0}$. The other class is related to the shape formation control, including $n_{\mathrm{cell}}$, $\kappa_{1}$, $\kappa_{2}$, $\kappa_{3}$ and $\kappa_{4}$.

	For the parameters adopted in the relative localization and shape localization agreement:

	\begin{itemize}
		\item The parameter $\Delta t$ represents the measurement time interval of the sensor, and $h$ represents the ratio of the time interval for collecting data to the measurement time interval of the sensor.
		The selections of $\Delta t$ and $h$ are generally related to the robot's motion speed and the IO measurement noise.
		In order to prevent the robot's motion displacement per unit time ($\Delta t$ multiplied by $h$) from being submerged in the measurement noise of the IO, we recommend that the norm of the robot's displacement per unit time be more than $10$ times the average measurement noise of the odometer.
		Below we provide a calculation example.
		Taking the Tb3 robot onboard odometer used in this work as an example, the odometer error (norm of the displacement measurement error) is generally less than $0.01$m within a short period of time (10 seconds) through our extensive practical verification.
		For the robot making localization-enhancement motion, its velocity is 
		mostly within the range of 0.05 and 0.22m/s (maximum velocity of Tb3 robot). 
		To ensure that the displacement of the robot per unit time ($\Delta t$ multiplied by $h$) is more than $10$ times the average noise level, the product of $\Delta t$ and $h$ is recommended to be around $2$. 
		As a result, the norm of the robot's displacement per unit time ($\Delta t$ multiplied by $h$) can be calculated roughly within the range of $0.1$ to $0.44$.
		
		\item The parameter $\lambda_{0}$ denotes the threshold for the ratio $\frac{\lambda_{\min}(S_{ij})}{\lambda_{\max}(S_{ij})}$ of the collected data matrix $S_{ij}$. 
		When $\frac{\lambda_{\min}(S_{ij})}{\lambda_{\max}(S_{ij})}$ is greater than $\lambda_{0}$, the localization-enhancement motion stop.
		$\lambda_{0}$ measures the amount of information collected, the larger $\frac{\lambda_{\min}(S_{ij})}{\lambda_{\max}(S_{ij})}$, the faster the relative position estimator converges. 
		For circular motion (radius between $0.3$m and $1$m) of robots at different angular velocities (between $0.1$m/s and $1$m/s), typically $\frac{\lambda_{\min}(S_{ij})}{\lambda_{\max}(S_{ij})}$ can reach around $0.2$.
		As a result, $\lambda_{0}$ is recommended to be around $0.1$ in practical applications to achieve faster convergence speed.
		\item The parameters $c_{1}$ and $\alpha$ are the gain and exponent of the seed robot relative position estimator, respectively. For $c_{1}$, generally it can be selected as a constant greater than $0$ but less than $10$. A large number of consensus related works are also selected in this way. 
		In our method, $\alpha$ is a parameter within the range of $0$ and $1$, as similar to most finite-time consensus results.
		In our method, it is shown that the shape agreement error $\tilde{q}$ will converge into an adjustable set $\Omega = \{ \lVert \tilde{q} \rVert \leq b \}$ in a finite time $t_{a} + t_{l}$ with the proposed estimation law, where $b$, $t_{a}$ and $t_{l}$ are defined as
		\begin{align}
			\label{b}
			b = \sqrt{\frac{ 2^{\frac{1+3\alpha}{2\alpha(\alpha+1)}} \left( \sum_{i=1}^{n} (|\mathcal{N}_{i}(t_{0})|\varepsilon)^{1+\alpha} \right)}{(1-\gamma) \lambda_{\min}(\mathcal{L} + \mathcal{B})^2}}
		\end{align}
		\begin{align}
			\label{t_l}
			t_{l} = \frac{2(1+\alpha)V_{l}(t_{a})^{\frac{1-\alpha}{2}}}{\lambda_{\min}(\mathcal{L} + \mathcal{B})\gamma(1-\alpha)}
		\end{align}
		\begin{align}
			\label{t_a}
			t_{a} = \mathrm{\max}\{ \frac{\mathrm{ln} (\varepsilon \lVert \tilde{p}_{ij,0}(t_{0}) \rVert^{-1})}{\mathrm{ln}(\lambda_{ij})} \Delta t, (i,j) \in \mathcal{G}(t_{0})\}
		\end{align}
		As far as the convergence time $t_{l}$ is concerned, it can be seen that the smaller $\alpha$, the smaller the convergence time $t_{l}$. 
		However, the smaller $\alpha$, the larger $b$, which means that the agreement error $\tilde{q}$ converges to a larger range in a faster time.
		So $\alpha$ should not be too small (approaching 0) nor too large (approaching 1).
		Through our extensive practical verification, $\alpha$ is recommended to be around $0.5$. 
		According to Eq. (18) in the revised manuscript, the smaller the value of $\alpha$, the faster the convergence speed of the estimator.
		\item The parameters $\delta_{t}$ and $\delta_{0}$ denote the unit time of the seed robot relative position estimator and threshold of estimator change rate, respectively. 
		For the seed robot relative position estimator, the rate $\frac{\lVert \hat{q}_{i,0}(t) - \hat{q}_{i,0}(t - \delta_t) \rVert}{\delta_t}$ reflects the changes in estimator values per unit time $\delta_{t}$. 
		When $\frac{\lVert \hat{q}_{i,0}(t) - \hat{q}_{i,0}(t - \delta_t) \rVert}{\delta_t}$ is greater than $\delta_{0}$, the agreement process can be considered as over.
		Generally, $\delta_{t0}$ can be easily chosen as $1$s, while for general precise requirements, $\delta_{0}$ can be chosen at around $0.01$, indicating that the estimator change rate is less than $1$ centimeters per second.
	\end{itemize}

	For the parameters adopted in the distributed shape formation control method:

	\begin{itemize}
		\item The parameter $n_{\mathrm{cell}}$ denotes the black cell number in the user-given binary image. The larger the value of $n_{\mathrm{cell}}$, the clearer the target image. 
		Generally, $40 \times 50$ image can be used to represent most complex shapes, so $n_{\mathrm{cell}}$ is recommended to be around $2000$.
		\item The parameters $\kappa_{1}$, $\kappa_{2}$, $\kappa_{3}$ and $\kappa_{4}$ denote the shape-entering command gain, shape-exploration command gain, collision-avoidance gain and velocity-alignment gain, respectively. 
		The larger $\kappa_{1}$, the greater the force that pushes the robot into shape. 
		The larger $\kappa_{2}$, the greater the force exerted by the robot in shape exploration. 
		The larger $\kappa_{3}$, the greater the collision avoidance force between robots. 
		The larger $\kappa_{4}$, the more obvious the speed alignment effect between robots. 
		Through our extensive practical verification, $\kappa_{1}$, $\kappa_{2}$ and $\kappa_{3}$ are chosen at around $10$, and $\kappa_{4}$ is recommended to be around $5$.
	\end{itemize} 

	The detailed parameters in our simulation and hand-ware experiments can be found in Table \ref{parameter}. From the parameter value in Table \ref{parameter}, it can be seen that our estimation algorithm and control algorithm do not make significant adjustments to parameters in simulations and hardware experiments with different target shapes and robot swarm scales. In other words, the algorithm proposed in this paper has low sensitivity to these parameters.

	\footnotesize
	\bibliographystyle{IEEEtran}
	\bibliography{IEEEabrv,references}
	
	
	\vfill
	
\end{document}